\documentclass[10pt,journal,compsoc]{IEEEtran}
\usepackage{amsmath}
\usepackage{amsfonts}
\usepackage{amssymb}
\usepackage{graphicx}
\usepackage{multirow}
\usepackage{subcaption}
\usepackage[ruled,boxed,noline]{algorithm2e}
\usepackage[font={small}]{caption}
\usepackage{enumerate}
\usepackage{color}
\usepackage{bbm}

\ifCLASSOPTIONcompsoc
  \usepackage[nocompress]{cite}
\else
  \usepackage{cite}
\fi
\ifCLASSINFOpdf
\else
\fi

\begin{document}
%
\title{An Automatic System for Unconstrained Video-Based Face Recognition}
%
%
%
%

\author{Jingxiao~Zheng,~
        Rajeev Ranjan,~
        Ching-Hui~Chen,~
        Jun-Cheng~Chen,~
        Carlos~D.~Castillo,~
        Rama~Chellappa
\IEEEcompsocitemizethanks{\IEEEcompsocthanksitem University of Maryland, College Park, MD.\protect\\
E-mail: jxzheng@umiacs.umd.edu}
}

\IEEEtitleabstractindextext{%
\begin{abstract}
Although deep learning approaches have achieved performance surpassing humans for still image-based face recognition, unconstrained video-based face recognition is still a challenging task due to large volume of data to be processed and intra/inter-video variations on pose, illumination, occlusion, scene, blur, video quality, etc. In this work, we consider challenging scenarios for unconstrained video-based face recognition from multiple-shot videos and surveillance videos with low-quality frames. To handle these problems, we propose a robust and efficient system for unconstrained video-based face recognition, which is composed of modules for face/fiducial detection, face association, and face recognition. First, we use multi-scale single-shot face detectors to efficiently localize faces in videos. The detected faces are then grouped respectively through carefully designed face association methods, especially for multi-shot videos. Finally, the faces are recognized by the proposed face matcher based on an unsupervised subspace learning approach and a subspace-to-subspace similarity metric. Extensive experiments on challenging video datasets, such as Multiple Biometric Grand Challenge (MBGC), Face and Ocular Challenge Series (FOCS), IARPA Janus Surveillance Video Benchmark (IJB-S) for low-quality surveillance videos and IARPA JANUS Benchmark B (IJB-B) for multiple-shot videos, demonstrate that the proposed system can accurately detect and associate faces from unconstrained videos and effectively learn robust and discriminative features for recognition.
\end{abstract}

}

\maketitle

\IEEEdisplaynontitleabstractindextext

%
\IEEEpeerreviewmaketitle

\section{Introduction}\label{sec:intro}

Face recognition is one of the most actively studied problems in computer vision and biometrics. Nowadays, video-based face recognition is an active research topic because of a wide range of applications including visual surveillance, access control, video content analysis, etc. Compared to still image-based face recognition, video-based face recognition is more challenging due to a much larger amount of data to be processed and significant intra/inter-class variations caused by motion blur, low video quality, occlusion, frequent scene changes, and unconstrained acquisition conditions. 

To develop the next generation of unconstrained video-based face recognition systems, two datasets have been recently introduced, IARPA Benchmark B (IJB-B) \cite{ijbb} and IARPA Janus Surveillance Video Benchmark (IJB-S) \cite{ijbs}, acquired under more challenging scenarios, compared to the Multiple Biometric Grand Challenge (MBGC) dataset \cite{MBGC} and the Face and Ocular Challenge Series (FOCS) dataset \cite{FOCS} which are collected in relatively controlled conditions. IJB-B and IJB-S datasets are captured in unconstrained settings and contain faces with much more intra/inter class variations on pose, illumination, occlusion, video quality, scale and etc.

\begin{figure}[t]
\centering
\includegraphics[width=0.95\linewidth]{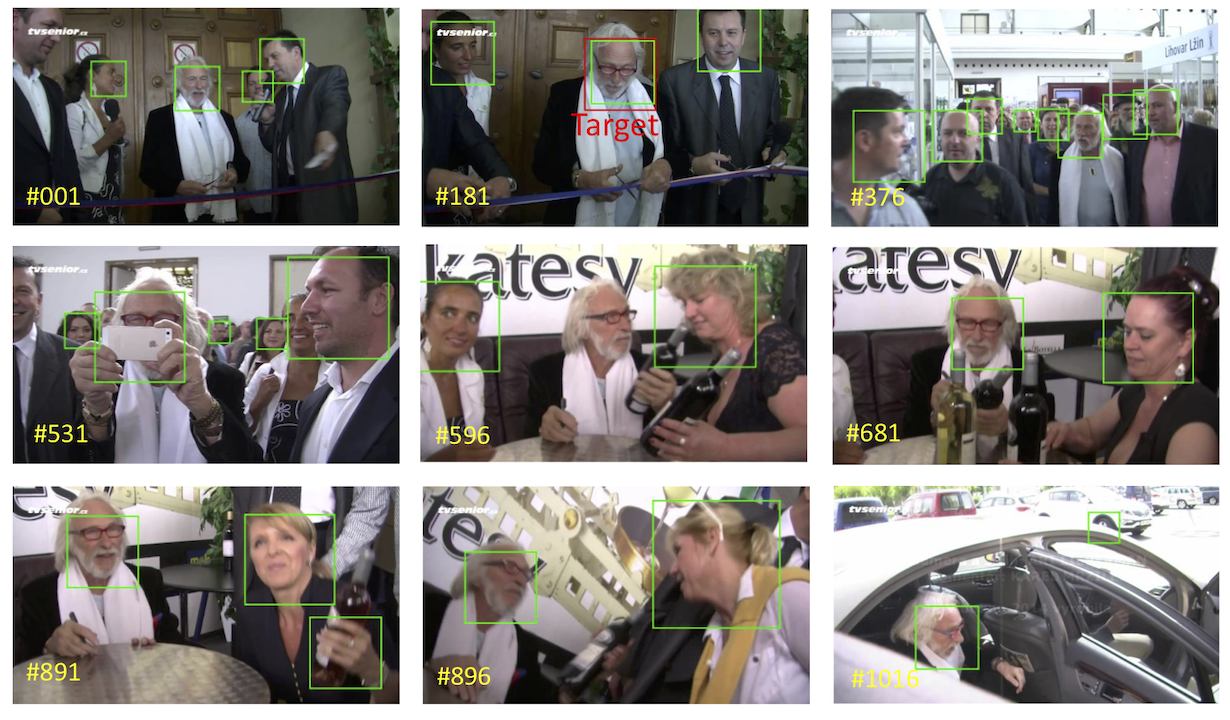}
\caption{Example frames of a multiple-shot probe video in the IJB-B dataset. The target annotation is in red box and face detection results from face detector are in green boxes.}
\label{fig:ijbb_example}
\end{figure}

The IJB-B dataset is a template-based dataset that contains 1845 subjects with 11,754 images, 55,025 frames and 7,011 videos where a template consists of a varying number of still images and video frames from different sources. These images and videos are collected from the Internet and are totally unconstrained, with large variations in pose, illumination, image quality etc. Samples from this dataset are shown in Figure~\ref{fig:ijbb_example}. In addition, the dataset comes with protocols for 1-to-1 template-based face verification, 1-to-N template-based open-set face identification, and 1-to-N open-set video face identification. For the video face identification protocol, the gallery is a set of still-image templates. The probe is a set of videos (e.g. news videos), each of which contains multiple shots with multiple people and one bounding box annotation to specify the subject of interest. Probes of videos are searched among galleries of still images. Since the videos are composed of multiple shots, it is challenging to detect and associate the faces for the subject of interest across shots due to large appearance changes. In addition, how to efficiently leverage information from multiple frames is another challenge, especially when the frames are noisy.

\begin{figure}[t]
\centering
\begin{subfigure}[b]{0.48\linewidth}
\centering
\includegraphics[width=\linewidth]{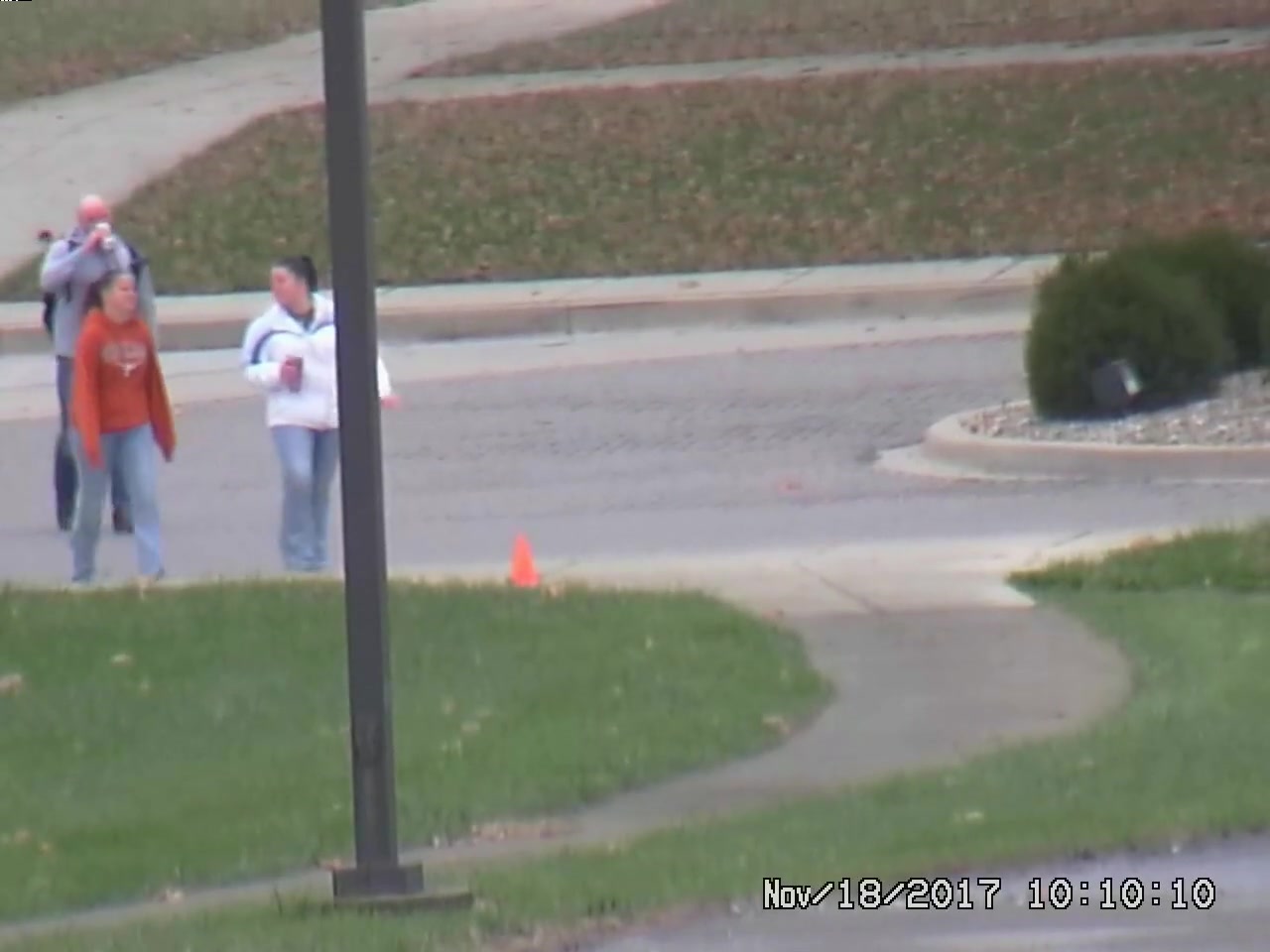}
\end{subfigure}
~
\begin{subfigure}[b]{0.48\linewidth}
\centering
\includegraphics[width=\linewidth]{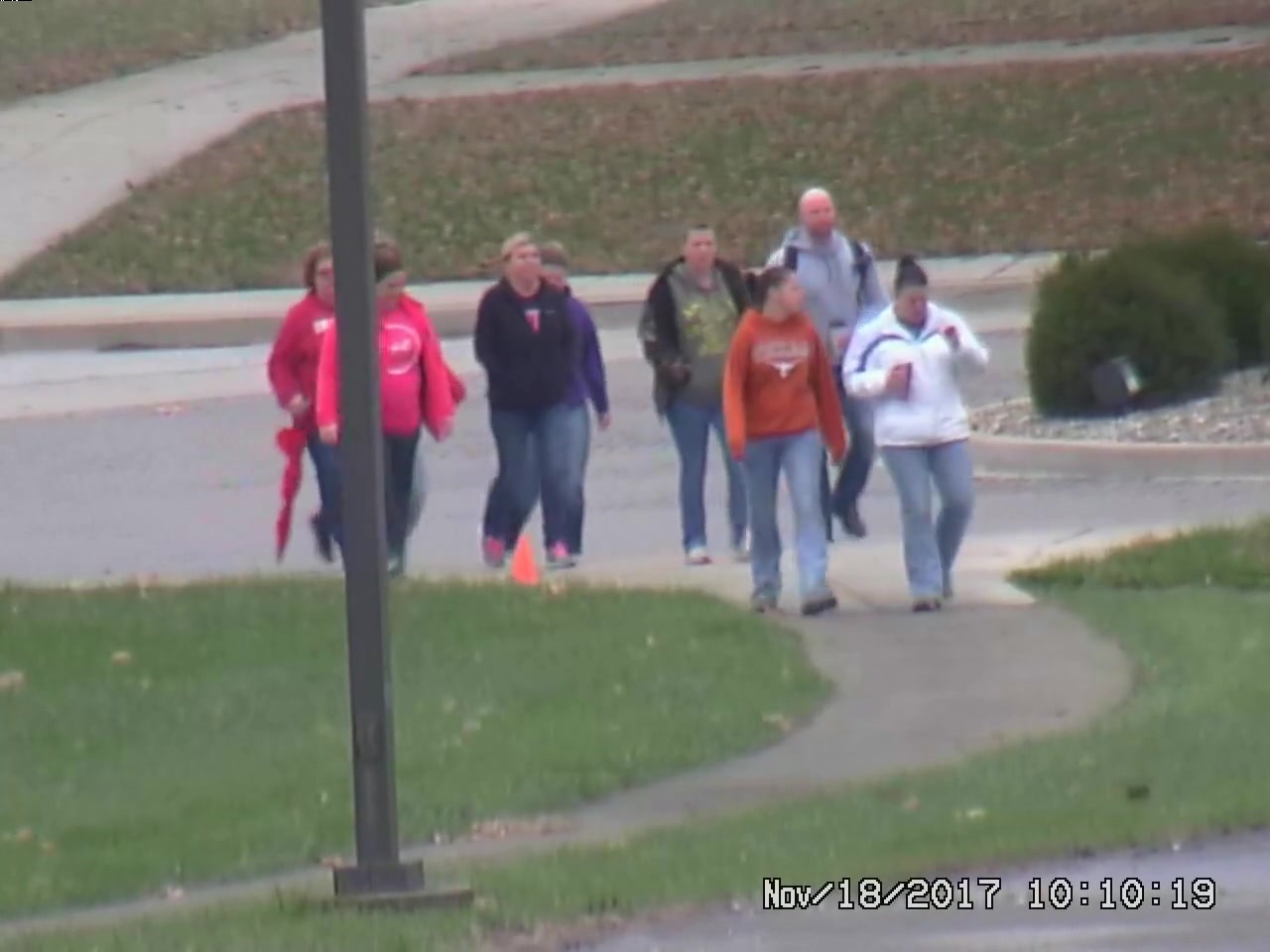}
\end{subfigure}
\\
\vspace{1em}
\begin{subfigure}[b]{0.48\linewidth}
\centering
\includegraphics[width=\linewidth]{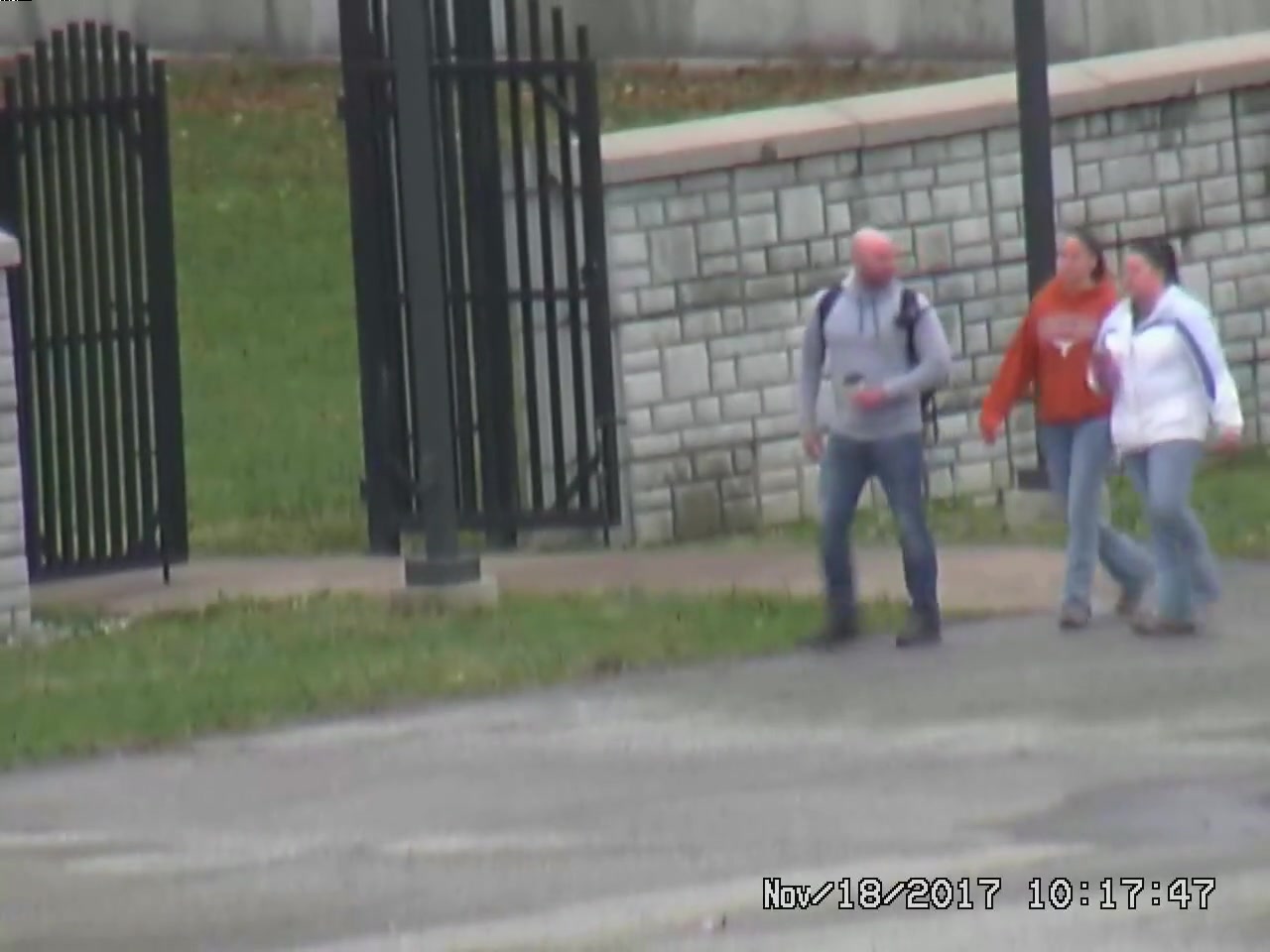}
\end{subfigure}
~
\begin{subfigure}[b]{0.48\linewidth}
\centering
\includegraphics[width=\linewidth]{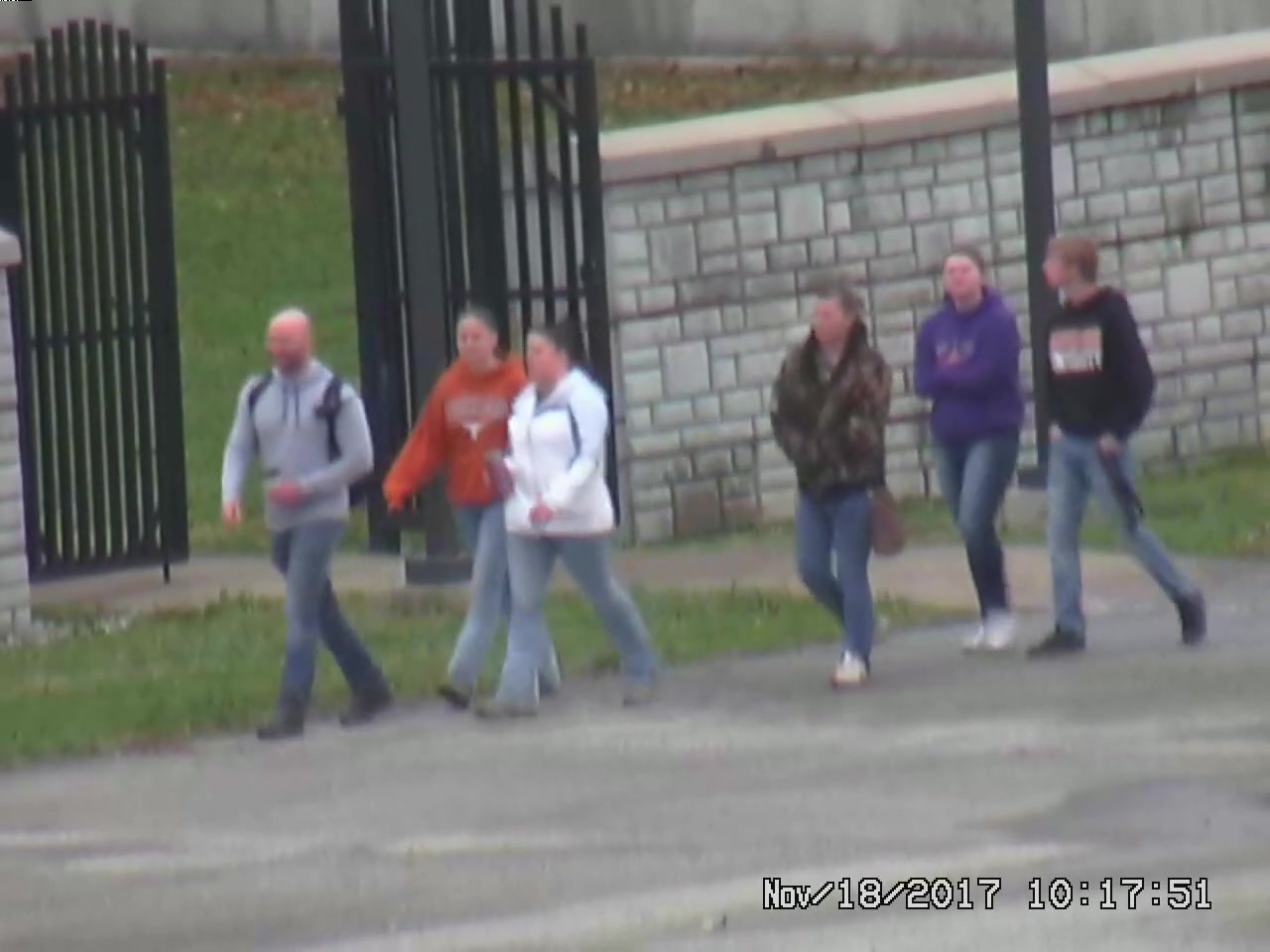}
\end{subfigure}
\caption{Example frames of two single-shot probe videos in the IJB-S dataset.}
\label{fig:cs6_example}
\end{figure}

Similar to the IJB-B dataset, the IJB-S dataset is also an unconstrained video dataset focusing on real world visual surveillance scenarios. It consists of 202 subjects from 1421 images and 398 surveillance videos, with 15,881,408 bounding box annotations. Samples of frames from IJB-S are shown in Figure~\ref{fig:cs6_example}. Three open-set identification protocols accompany this dataset for surveillance video-based face recognition where each video in these protocols is captured from a static surveillance camera and contains single or multiple subjects: (1) in surveillance-to-single protocol, probes collected from surveillance videos are searched in galleries consisting of one single high-resolution still image; (2) in surveillance-to-booking protocol, same probes are searched among galleries consisting of seven high-resolution still face images covering frontal and profile poses. Probe templates in (1) and (2) should be detected and constructed by the recognition system itself; (3) in the most challenging surveillance-to-surveillance protocol, both gallery and probe templates are from videos, which implies that probe templates need to be compared with relatively low quality gallery templates.

From these datasets, we summarize the four common challenges in video-based face recognition as follows:
\begin{enumerate}
\item For video-based face recognition, test data are from videos where each video contains tens of thousands of frames and each frame may have several faces. This makes the scalability of video-based face recognition a challenging problem. In order to make the face recognition system to be operationally effective, each component of the system should be fast, especially face detection, which is often the bottleneck in recognition.

\item Since faces are mostly from unconstrained videos, they have significant variations in pose, expression, illumination, blur, occlusion and video quality. Thus, any face representations we design must be robust to these variations and to errors in face detection and association steps.

\item Faces with same identity across different video frames need to be grouped by a reliable face association method. Face recognition performance will degrade if faces with different identities are grouped together. Videos in the IJB-B dataset are acquired from multiple shots involving scene and view changes, while most videos in IJB-S are low-quality remote surveillance videos. These conditions increase the difficulty of face association. 

\item Since each video contains different number of faces for each identity, the next challenge is how to efficiently aggregate a varying-length set of features from the same identity into a fixed-size or unified representation. Exploiting the correlation information in a set of faces generally results in better performance than using only a single face.
\end{enumerate}

\begin{figure*}
\centering
\includegraphics[width=0.95\textwidth]{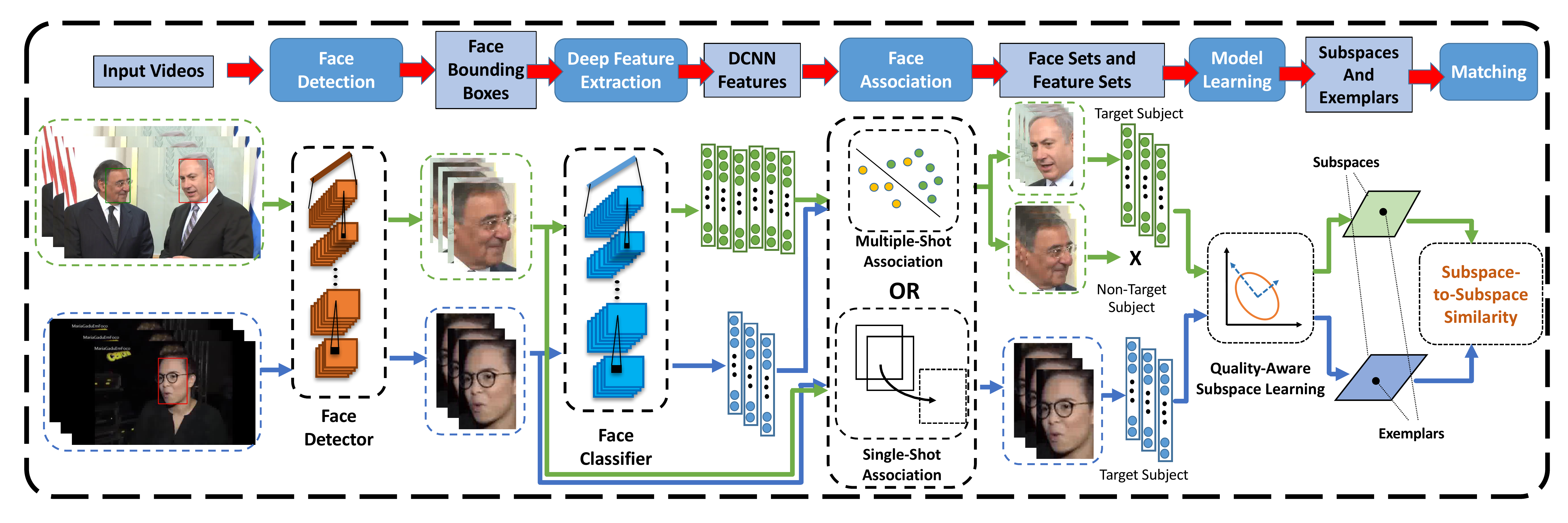}
\caption{Overview of the proposed system.}
\label{fig:overview}
\end{figure*}

Recently, with the availability of powerful GPUs and large amounts of labeled training data, deep convolutional neural networks (DCNNs) have demonstrated impressive performances for many computer vision tasks such as object recognition \cite{He2015residual}, object detection \cite{fasterrcnn} and semantic segmentation \cite{deeplab}. DCNNs have also produced state-of-the-art results for face detection and still face-based recognition tasks as reported in \cite{allinone}, \cite{ssh}, \cite{Taigman2014}, \cite{Parkhi15}, \cite{Schroff2015}, \cite{JC15}. Thus, by utilizing the power of deep networks, we could detect faces more accurately from videos and produce robust deep representations for these faces than traditional approaches. However, the speed of face detectors is still a critical bottleneck in the face recognition pipeline. Single Shot MultiBox Detector (SSD) \cite{ssd} and YOLO \cite{yolo} provide a solution for fast-speed face detector. These one-step face detectors achieve comparable results to state-of-the-art two-step detectors \cite{fasterrcnn} \cite{maskrcnn}, but are several times faster. Also, instead of using the traditional cross-entropy loss, Ranjan \emph{et al.} \cite{crystal} recently introduced the crystal loss, which functions by constraining the deep features to lie on a hypersphere, and achieves impressive result on the challenging face recognition dataset, IARPA JANUS Benchmark C (IJB-C) \cite{ijbc}. 

Many tracking techniques \cite{trackparticle}, \cite{robustmft}, \cite{OnlineMD}, \cite{SORT} have been proposed to associate face images of a subject in videos containing a single shot, by utilizing spatial, temporal, and appearance affinity. Simple Online and Realtime Tracking (SORT) \cite{SORT} is a very fast and efficient tracker for tracking multi-target bounding boxes. However, SORT is not effective for multiple-shot videos where frequent scene changes are present across shots. To address the large variations across shots, we use a face association method, Target Face Association (TFA) proposed in \cite{TFA}, which adaptively updates face representations by one-shot SVM kernel to retrieve a set of representative face images in a multiple-shot video for video-based face recognition. More details are described in Section~\ref{faceas}.

For feature aggregation in video-based face recognition, temporal deep learning model such as Recurrent Neural Network (RNN) can be applied to yield a fixed-size encoded face representation. However, large-scale labeled training data is needed to learn robust representations, which is very expensive to collect in the context of video-based recognition problem. This is also true for the adaptive pooling method \cite{qan}, \cite{nan} for image set-based face recognition problem. For IJB-B and IJB-S datasets, lack of large-scale training data makes it impossible to train an RNN-based method. On the contrary, representative and discriminative models based on manifolds and subspaces have also received attention for image set-based face recognition \cite{Wang2012} \cite{wang2009}. These methods model sets of image samples as manifolds or subspaces and use appropriate similarity metric for set-based identification and verification. One of the main advantages of subspace-based methods is that different from sample mean, the subspace representation encodes the correlation information between samples. In low-quality videos, faces have significant variations due to blur, extreme poses and low resolution. Exploiting correlation between samples by subspacess will help learn a more robust representation to capture these variations. Also, a fixed-size representation can be learned from an arbitrary number of images or video frames.

To summarize, we propose an automatic system by integrating deep learning components to overcome the challenges in unconstrained video-based face recognition. The proposed system first detects faces and facial landmarks using state-of-the-art DCNN face detectors. Single Shot Detector (SSD) for faces \cite{icipssd} and Deep Pyramid Single Shot Face Detector (DPSSD) \cite{systbiom} are used for different scenarios. The former is an efficient face detector which is effective for videos collected from the Internet where faces are relatively larger than the ones in surveillance videos. The latter is a multi-scale face detector that can produce reliable and accurate face detections at different scales, thus is capable of detecting tiny and blurred faces which are common in surveillance videos. Both detectors are fast, which can tackle the first challenge of processing speed. Next, we extract deep features from the detected faces using state-of-the-art DCNNs for face recognition. We use crystal loss to train the DCNN network, to address the second challenge of generating a robust representation. Besides SORT for single-shot videos, TFA is used to cluster target faces from multiple-shot videos. By incorporating SORT and TFA into our system, together with robust face detectors, face SSD and DPSSD, our system overcomes the third challenge of video-based face detection and association. Finally, in the proposed face recognition system, we learn a subspace representation from each video template and match pairs of templates using principal angles-based subspace-to-subspace similarity measure on the learned subspace representations. This helps us to handle the fourth challenge of representation aggregation. An overview of the proposed system is shown in Figure~\ref{fig:overview}.

We evaluate our face recognition system on the challenging IJB-B and IJB-S datasets, as well as MBGC and FOCS datasets, and the results demonstrate that the proposed system achieves improved performance over other deep learning-based baselines and state-of-the-art approaches.

The main contributions of the proposed system are summarized as
follows:

\begin{itemize}
\item We propose an automatic video-based face recognition system with components including face/fiducial detection, face association, and face recognition.
\item We propose a quality-aware subspace learning approach for face feature aggregation.
\item We compute the video template-to-template similarity using a subspace-to-subspace similarity metric for video-based face recognition. A quality-aware subspace-to-subspace similarity metric is also proposed.
\end{itemize}

The rest of this paper is organized as follows: in Section~\ref{sec:related}, we briefly review some related works. In Section~\ref{sec:method}, we introduce the proposed system in detail. In Section~\ref{sec:exp}, we discuss the implementation details and present the experimental results on four datasets. Finally, conclusions are presented in Section~\ref{sec:conclusion}.

\section{Related Work}\label{sec:related}

\textbf{1. Deep Learning for Face Recognition:}
Taigman \emph{et al.} \cite{Taigman2014} learned a DCNN model on the frontalized faces generated from 3D shape models built from face dataset. Sun \emph{et al.} \cite{Sun2014} \cite{Sun2014Sparse} achieved results surpassing human performance for face verification on the LFW dataset \cite{LFW}. 
Schroff \emph{et al.} \cite{Schroff2015} adopted the GoogLeNet trained for object recognition to face recognition and trained on a large-scale unaligned face dataset. Parkhi \emph{et al.} \cite{Parkhi15} achieved impressive results using a very deep convolutional network based on VGGNet for face verification. Ding \emph{et al.} \cite{Ding2016} proposed a trunk-branch ensemble CNN model for video-based face recognition. Chen \emph{et al.} \cite{JC15} trained a 10-layer CNN on CASIAWebFace dataset \cite{casia} followed by a joint Bayesian metric and achieved state-of-the-art performance on the IJB-A \cite{IJBA} dataset. Chen \emph{et al.} \cite{jc_ijcv} further extended \cite{JC15} and designed an end-to-end system for unconstrained face recognition and reported very good performance on IJB-A, JANUS CS2, LFW and YouTubeFaces \cite{YTF} datasets. In order to tackle the training bottleneck for face recognition network, Ranjan \emph{et al.} \cite{crystal} proposed the crystal loss to train the network on very large scale training data. It achieved state-of-the-art result on the challenging IJB-C dataset for unconstrained face recognition. 

\textbf{2. Face Detection:}
Najibi \emph{et al.} \cite{ssh} proposed a single-stage fully convolutional network for face detection. It is fast and achieves state-of-the-art results on WIDER \cite{WIDER}, FDDB \cite{FDDB} and Pascal-Faces \cite{PascalFace} datasets. HyperFace \cite{hyperface} is the first multi-task network that can simultaneously detect faces, extract fiducials, estimate pose and recognize gender. In \cite{allinone}, Ranjan \emph{et al.} built on \cite{hyperface} and used multi-task learning to simultaneously obtain face-related information like face bounding boxes, fiducials, facial pose, gender and identity. Chen \emph{et al.} \cite{icipssd} proposed a multi-task face detector based on the single stage SSD detector with extra branches. It achieved competitive performance on FDDB, AFW and PASCAL-Faces datasets with the similar speed as SSD.

\textbf{3. Face Retrieval and Tracking:}
In the literature on face retrieval, many methods have been proposed \cite{SwamiBTAS}, \cite{Ortiz}, \cite{Cinbis}. Sivic et al. \cite{Sivic} proposed a method that retrieves the target subject using a set of images that contains extensive variations of exemplars. Arandjelovic \emph{et al.} \cite{Arandjelovic} proposed an end-to-end video face retrieval system with several processing steps.

There are also many works on face tracking and association. Zhou \emph{et al.} \cite{trackparticle} incorporated appearance-based models in a particle filter to realize robust visual tracking. \cite{robustmft} proposed a multi-pose face tracking approach in two stages using multiple cues. Comaschi \emph{et al.} \cite{OnlineMD} also proposed an online multi-face tracker using detector confidence and a structured SVM. An efficient tracker, SORT \cite{SORT}, achieves comparable results to other state-of-the-art methods with 20 times faster speed. Du \emph{et al.} \cite{du_association} proposed a conditional random field (CRF) framework to associate faces by utilizing the similarity of facial appearance, location, motion, and body appearance.


\textbf{4. Image Set/Video-based Recognition:}
For image set-based recognition, Wang \emph{et al.} \cite{Wang2012} proposed a Manifold-to-Manifold Distance (MMD) for face recognition based on image set. In \cite{cov}, the proposed approach models the image set with its second-order statistic for image set classification. Chen \emph{et al.} \cite{DFRV_2012} and \cite{DFRV_2015} proposed a video-based face recognition algorithm using sparse representation and dictionary learning. Zheng \emph{et al.} \cite{hybrid} proposed a hybrid dictionary learning and matching approach for video-based face recognition.


\section{Proposed Method}\label{sec:method}

 For each video, we first detect faces from video frames and align them using the detected fiducial points. Deep features are then extracted for each detected face using our DCNN models for face recognition. Based on different scenarios, we use face association or face tracking to construct face templates with unique identities. For videos with multiple shots, we use the proposed face association technique TFA \cite{TFA} to collect faces from the same identities across shots. For single-shot videos, we use the face tracking algorithm SORT introduced in \cite{SORT} to produce tracklets of faces.
After templates are constructed, in order to aggregate the representation of videos, subspaces are learned using quality-aware principal component analysis. Subspaces along with quality-aware exemplars of templates are used to produce the similarity scores between video pairs by a quality-aware principal angle-based subspace-to-subspace similarity measure.
In the following sections, we discuss the proposed video-based face recognition system in details.

\subsection{Face/Fiducial Detection}
The first step in our face recognition pipeline is to detect faces in images (usually for galleries) and videos. The challenges mostly come from the low quality of surveillance videos. Our detector should balance the precision and recall of bounding boxes since we do not want too many non-faces to be input to the rest of the pipeline. Also we do not want our detector to miss too many faces since they can provide useful information for recognition. Thus we use two DCNN-based detectors in our pipeline based on different distributions of input.

\subsubsection{SSD Face Detector}
For regular images and video frames, faces are relatively bigger and with higher resolution. We use SSD trained with the WIDER face dataset as our face detector \cite{icipssd}. 

\subsubsection{DPSSD Face Detector}
For small and remote faces in surveillance videos, because of the domain difference, a traditional face detector cannot perform well on detecting these tiny faces. Thus for surveillance videos, we use a novel DCNN-based face detector, called Deep Pyramid Single Shot Face Detector (DPSSD) \cite{systbiom} for face detection. It is fast and capable of detecting tiny faces, which is very suitable for face detection in videos.

After raw face detection bounding boxes are generated using either SSD or DPSSD detectors, we use All-in-One Face \cite{allinone} for fiducial localization. It is followed by a seven-point face alignment step based on the similarity transform on all the detected faces.

\subsection{Deep Feature Representation}
After faces are detected and aligned, we use the DCNN models to represent each detected face. The models are state-of-the-art networks with different architectures for face recognition. Different architectures provide different error patterns during testing. After fusing the results from different models, we achieve performance better than a single model. The overview of these networks along with their training details are described in Section~\ref{sec:imp}.

\subsection{Face Association}\label{faceas}
In previous steps, we obtain raw face detection bounding boxes using our detectors. Features for the detected bounding boxes are extracted using face recognition networks. Since video frames are of relatively low quality, the detector outputs will consist of both faces and non-face objects. Even for correct detection, faces from different identities will appear in the same video containing multiple subjects. Many popular video-based face recognition datasets have testing protocols in which faces are matched in templates (aka image sets) instead of single images. If the constructed templates are not clean, outliers in the template will have negative influence on the discriminative power of the face representation and adversely affect the face recognition performance. Thus, the next important step in our face recognition pipeline is to combine the detected bounding boxes from the same identity to construct templates for good face recognition result.

\subsubsection{Face Association for Single-Shot Videos}

For single-shot videos, which means the bounding boxes of a certain identity will probably be contiguous, we rely on SORT \cite{SORT} to build tracklets for each identity. 
SORT is a real-time online tracking algorithm which approximates the dynamics with linear Gaussian state space models and associates detection bounding boxes in every frame using Kalman Filters.

\subsubsection{Face Association for Multi-Shot Videos}\label{sec:TFA}
For multi-shot videos, it is challenging to continue tracking across different scenes. Thus, face association based on the appearance (deep representation in our case) of the face bounding boxes instead of their position is more robust to scene changes in these videos. In the proposed system, we use \cite{TFA} to adaptively update the face representation through one-shot SVM. The details are described below.

\textbf{a) Face Pre-Association by Tracking}
Starting from the annotated face, an off-the-shelf tracking technique \cite{ofstracker} is first applied to track the target face, which is called pre-association, to collect subsequent faces of high quality. Faces in the same tracklet are utilized as the initial positive training set.

A tracklet is built by associating the face detection bounding box that has the highest IoU ratio with the tracking bounding box in each frame. We only track faces in the first k frames, or stop at scene boundaries, since tracking will be unreliable if occlusion, motion and scene changes occur.

\textbf{b) One-Shot SVM Learning}
In a video, suppose the annotated target face is indicated by a bounding box $b_0$ in frame $f_0$. There are a total of $m$ bounding boxes discovered by the detector. These face bounding boxes are denoted as $b_1, b_2,\dots, b_m$, which are present in frames $f_1, f_2, \dots, f_m$, respectively. The feature corresponding to the face bounding box $b_i$ is denoted as $\mathbf{x}_i$. Given the initial positive training set, a subject-specific linear SVM is trained to establish the intra/inter-shot face association of the target face.

The index set of initial positive instances is represented as $S_p = \{0\}\cup T$, where $T$ are indices of pre-associated bounding boxes. The negative training instances can be discovered by the cannot-link relation. The cannot-link relation between the $i$th and $j$th bounding boxes is defined as
\vspace{-5pt}
\begin{align}
g_{i,j} = \begin{cases}
1 & \text{if } r_{i,j} \leq \gamma,\;f_i = f_j,\;i, j \in \{0, 1, \dots , m\}\\
0 & \text{otherwise}
\end{cases}
\end{align}
\vspace{-5pt}

where $r_{i,j}$ is the IoU ratio between bounding box $b_i$ and $b_j$, $\gamma$ is the corresponding IoU threshold. Thus, $g_{i,j} = 1$ indicates that the $i$th and $j$th bounding boxes appear in the same frame, cannot belong to the same face, and should not be identified as the same subject. Therefore the index set of within-video negative instances is represented as $S_n = \cup_{j\in S_p}\{i |g_{i,j} = 1\}$.

If there is no within-video training instance, a background negative set $\{\mathbf{x}_i\}_{i=m+1}^{m+l}$ with size $l$ collected from an external face dataset will be used with corresponding index set $S_b = \{m + 1,m + 2, \dots, m + l\}$.

Then, the linear SVM is trained by the combined training data $\{(\mathbf{x}_i, y_i)|i \in (S_p \cup S_n \cup S_b)\}$, where the data label is defined as
\begin{equation}
y_i = \begin{cases}
1 & \text{if } i \in S_p,\\
-1 & \text{otherwise.}
\end{cases}
\end{equation}

\begin{figure}[t]
\centering
\includegraphics[width=0.75\linewidth]{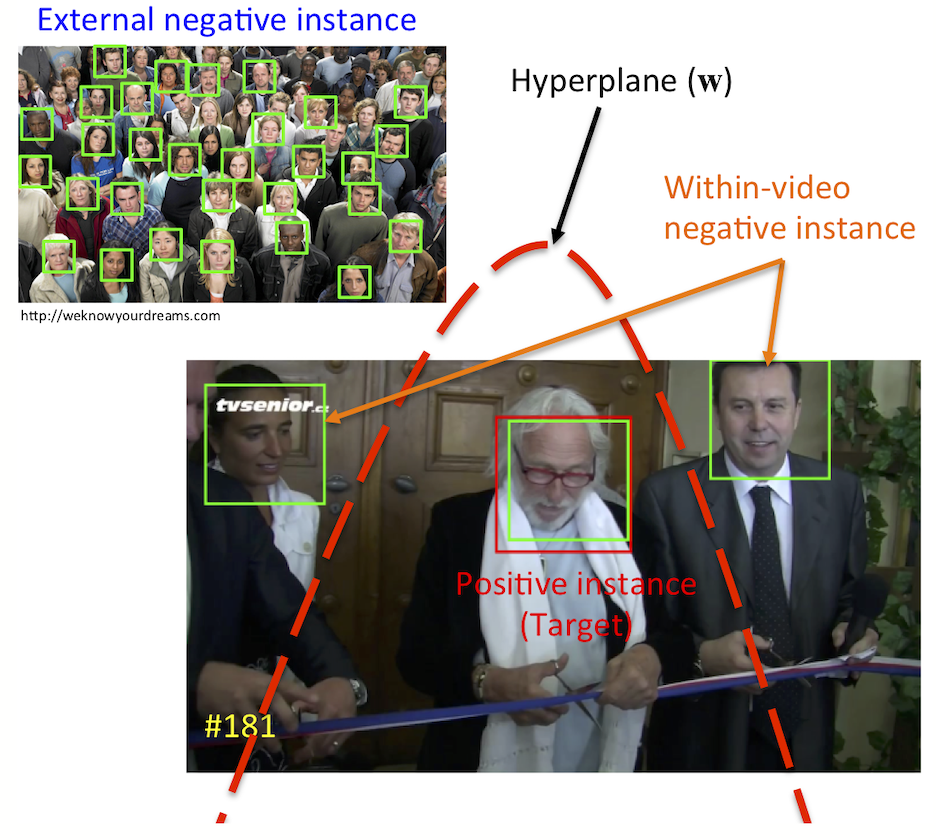}
\caption{Example of training instances for the SVM.}
\label{fig:TFA}
\end{figure}

The weight vector $\mathbf{w}$ of the linear SVM is solved using the max-margin framework
\begin{align}\label{svm}
\operatorname*{minimize}_{\mathbf{w}}&\frac{1}{2}\mathbf{w}^T\mathbf{w} + C_p\sum_{i\in S_p}\max[0, 1-y_i\mathbf{w}^T\bar{\mathbf{x}}_i]^2\nonumber\\
&+ \mathbbm{1}[S_n\neq\varnothing]C_n\sum_{i\in S_n}\max[0, 1-y_i\mathbf{w}^T\bar{\mathbf{x}}_i]^2\nonumber\\
&+ \mathbbm{1}[S_b\neq\varnothing]C_b\sum_{i\in S_b}\max[0, 1-y_i\mathbf{w}^T\bar{\mathbf{x}}_i]^2
\end{align}

where $C_p$, $C_n$ and $C_b$ are weights related to the number of training samples, $\bar{\mathbf{x}}_i=\begin{bmatrix}\mathbf{x}_i^T/\|\mathbf{x}_i\|, 1\end{bmatrix}^T$ is the normalized feature, $\mathbbm{1}[\cdot]$ is the indicator function. An illustration of the training instances is shown in Figure~\ref{fig:TFA}.

After the SVM is learned, faces that are classified as positive will be regarded as the associated face set to the target subject as
{
\begin{equation}
A = \{0\}\cup\{i|\mathbf{w}^T\mathbf{x}_i > 0, i=1,2,\dots,m\}
\end{equation}
}
Since faces in set $A$ may come from the same frame, $A$ is further refined by iteratively remove the least likely
instance among those instances that violate the cannot-link constraints.

Subsequently, face assignment and refinement procedure is applied on every annotated bounding box, and thus we obtain the associated face set for each annotation. 

\subsection{Model Learning: Deep Subspace Representation}\label{Subspace}

Once deep features are extracted for each face template, since each template contains a varying number of faces, these features need to be further aggregated into a fixed-size or a unified representation for efficient face recognition.

The simplest representation of a set of samples is the sample mean. However, this video template contains faces with different quality and large variations in illumination, blur and pose. Since the average treats all the samples equally, the outliers may deteriorate the discriminative power of the representation. As compared to other feature aggregation approaches that require a large amount of extra training data which are not available for datasets like IJB-B and IJB-S, we propose to represent face templates by subspaces.

\subsubsection{Subspace Learning from Deep Features}
A $d$-dimensional subspace $S$ can be uniquely defined by a set of orthonormal bases $\mathbf{P}\in\mathbb{R}^{D\times d}$, where $D$ is the dimensionality of features. Given face features from a video sequence $\mathbf{Y}\in\mathbb{R}^{D\times N}$, where $N$ is the sequence length, $\mathbf{P}$ can be found by optimizing:
\begin{align}\label{PCA}
\operatorname*{minimize}_{\mathbf{P}, \mathbf{X}}\;\|\mathbf{Y} - \mathbf{P}\mathbf{X}\|_F^2\quad s.t.\;\mathbf{P}^T\mathbf{P} = \mathbf{I}
\end{align}
which is the reconstruction error of features $\mathbf{Y}$ in the subspace $S$. It is exactly the principal component analysis (PCA) problem and can be easily solved by eigenvalue decomposition. Let $\mathbf{Y}\mathbf{Y}^T = \mathbf{U}\boldsymbol{\Lambda}\mathbf{U}^T$ be the eigenvalue decomposition, where $\mathbf{U} = \begin{bmatrix}\mathbf{u}_1, \mathbf{u}_2,\cdots,\mathbf{u}_D\end{bmatrix}$ are eigenvectors and $\boldsymbol{\Lambda} = diag\{\lambda_1, \lambda_2, \dots, \lambda_D\}$ with $\lambda_1\geq\lambda_2\geq\dots\geq\lambda_D$ are the corresponding eigenvalues, we have $\mathbf{P} = \begin{bmatrix}\mathbf{u}_1, \mathbf{u}_2,\cdots,\mathbf{u}_d\end{bmatrix}$ which consists of the first $d$ basis in $\mathbf{U}$. We use \textbf{Sub} to denote this basic subspace learning algorithm.

\subsubsection{Quality-Aware Subspace Learning from Deep Features}
In a face template from videos, faces contain large variations in pose, illumination, occlusion, etc. Even in a tracklet, faces have different poses because of the head movement, or being occluded at some frames because of the interaction with the environment. When learning the subspace, treating the frames equally is not an optimal solution. In our system, the detection score for each face bounding box given by the face detector can be used as a good indicator of the quality of faces, as shown in \cite{crystal}. Hence, following the quality pooling proposed in \cite{crystal}, we propose quality-aware subspace learning based on detection scores. The learning problem modifies \eqref{PCA} as
\begin{align}
\operatorname*{minimize}_{\mathbf{P}, \mathbf{X}}\;\sum_{i=1}^{N}\tilde{d}_i\|\mathbf{y}_i - \mathbf{P}\mathbf{x}_i\|_2^2\quad s.t.\;\mathbf{P}^T\mathbf{P} = \mathbf{I}
\end{align}
where $\tilde{d}_i=softmax(q l_i)$ is the normalized detection score of face $i$, $q$ is the temperature parameter and
\begin{equation}\label{li}
l_i = \min(\frac{1}{2}\log\frac{d_i}{1-d_i}, t)
\end{equation}
which is upper bounded by threshold $t$ to avoid extreme values when the detection score is close to 1.

Let $\tilde{\mathbf{Y}} = \begin{bmatrix}\sqrt{d_1}\mathbf{y}_1,\cdots,\sqrt{d_N}\mathbf{y}_N\end{bmatrix}$ be the normalized feature set, and the corresponding eigenvalue decomposition to be $\tilde{\mathbf{Y}}\tilde{\mathbf{Y}}^T = \tilde{\mathbf{U}}\tilde{\boldsymbol{\Lambda}}\tilde{\mathbf{U}}^T$.
We have
\begin{equation}\label{PD}
\mathbf{P}_D = \begin{bmatrix}\tilde{\mathbf{u}}_1, \tilde{\mathbf{u}}_2,\cdots,\tilde{\mathbf{u}}_d\end{bmatrix}
\end{equation}
which consists of the first $d$ bases in $\tilde{\mathbf{U}}$. The new subspace is therefore learned by weighting samples differently according to their quality. This quality-aware learning algorithm is denoted as \textbf{QSub}.
\subsection{Matching: Subspace-to-Subspace Similarity for Videos}

After deep subspace face representations are learned for video templates, inspired by manifold-to-manifold distance \cite{Wang2012}, we measure the similarity between two video templates of faces using a subspace-to-subspace similarity measure. In this part, we first introduce the widely used measure based on principal angles. Then we propose several weighted subspace-to-subspace measures which take the importance of bases into consideration.

\subsubsection{Principal Angles and Projection Metric}
One of the mostly used subspace-to-subspace similarity is based on principal angles. The principal angles $0\leq \theta_1\leq\theta_2\leq\dots\leq\theta_r\leq \frac{\pi}{2}$ between two linear subspaces $S_1$ and $S_2$ can be computed by Singular Value Decomposition (SVD).

Let $\mathbf{P}_1\in\mathbb{R}^{D\times d_1}$, $\mathbf{P}_2\in\mathbb{R}^{D\times d_2}$, denoting the orthonormal basis of $S_1$ and $S_2$, respectively. The SVD of $\mathbf{P}_1^T\mathbf{P}_2$ is $\mathbf{P}_1^T\mathbf{P}_2 = \mathbf{Q}_{12}\boldsymbol{\Lambda}\mathbf{Q}_{21}^T$, where $\boldsymbol{\Lambda} = diag\{\sigma_1,\sigma_2,\dots,\sigma_r\}$. $\mathbf{Q}_{12}$ and $\mathbf{Q}_{21}$ are orthonormal matrices. The singular
values $\sigma_1,\sigma_2,\dots,\sigma_r$ are exactly the cosine of the principal angles as $\cos\theta_k = \sigma_k,\; k=1, 2, \dots, r$.

Projection metric \cite{PM} is a popular similarity measure based on principal angles:
\begin{equation}
s_{PM}(S_1,S_2)=\sqrt{\frac{1}{r}\sum_{k=1}^r\cos^2\theta_k}
\end{equation}

Since $\|\mathbf{P}_1^T\mathbf{P}_2\|_F^2 = \|\mathbf{Q}_{12}\boldsymbol{\Lambda}\mathbf{Q}_{21}^T\|_F^2=\|\boldsymbol{\Lambda}\|_F^2=\sum_{k=1}^r\sigma_k^2=\sum_{k=1}^r\cos^2\theta_k$, we have
\begin{equation}\label{PM}
s_{PM}(S_1,S_2) = s_{PM}(\mathbf{P}_1,\mathbf{P}_2)=\sqrt{\frac{1}{r}\|\mathbf{P}_1^T\mathbf{P}_2\|_F^2}
\end{equation}
and there is no need to compute the SVD explicitly. We use \textbf{PM} to denote this similarity measure.

\subsubsection{Exemplars and Basic Subspace-to-Subspace Similarity}
Existing face recognition systems usually use cosine distance between exemplars to measure the similarity between templates. Exemplar of a template is defined as its sample mean, as $\mathbf{e}=\frac{1}{L}\sum_{i=1}^{L}\mathbf{y}_{i}$,
where $\mathbf{y}_i$ are samples in the template. Exemplars mainly capture the average and global representation of the template. On the other hand, the projection metric we introduced above measures the similarity between two subspaces, which models the correlation between samples. Hence, in the proposed system, we make use of both of them by fusing their similarity scores as the subspace-to-subspace similarity between two video sequences.

Suppose subspaces $\mathbf{P}_1\in\mathbb{R}^{D\times
d_1}$ and $\mathbf{P}_2\in\mathbb{R}^{D\times d_2}$ are learned from a pair of video templates $\mathbf{Y}_1\in\mathbb{R}^{D\times L_1}$ and
$\mathbf{Y}_2\in\mathbb{R}^{D\times L_2}$ in deep features respectively, by either \textbf{Sub} or \textbf{QSub} methods introduced in Section~\ref{Subspace}. Their exemplars are
$\mathbf{e}_1=\frac{1}{L_1}\sum_{i=1}^{L_1}\mathbf{y}_{1i}$ and
$\mathbf{e}_2=\frac{1}{L_2}\sum_{i=1}^{L_2}\mathbf{y}_{2i}$ respectively. Combining the orthonormal bases and exemplars, the subspace-to-subspace similarity can be computed as:
\begin{align}
s(\mathbf{Y}_1,\mathbf{Y}_2)&=s_{Cos}(\mathbf{Y}_1,\mathbf{Y}_2)+\lambda s_{PM}(\mathbf{P}_1, \mathbf{P}_2)\nonumber\\
&=\frac{\mathbf{e}_1^T\mathbf{e}_2}{\|\mathbf{e}_1\|_2\|\mathbf{e}_2\|_2} + \lambda\sqrt{\frac{1}{r}\|\mathbf{P}_1^T\mathbf{P}_2\|_F^2}
\end{align}
where $s_{Cos}(\mathbf{Y}_1,\mathbf{Y}_2)$
is the cosine distance between exemplars, denoted as \textbf{Cos}, and $s_{PM}(\mathbf{P}_1, \mathbf{P}_2)$ is computed by \eqref{PM}. Since the DCNN features are more robust if we keep their signs, instead of using $s_{Cos}^2(\mathbf{Y}_1,\mathbf{Y}_2)$ as in \cite{Wang2012} where the sign information is lost, we use $s_{Cos}(\mathbf{Y}_1,\mathbf{Y}_2)$ in our formulation. Accordingly, we also take the square root of the principal angle term to keep the scale consistent. $\lambda$ here is a hyperparameter that balances the cosine similarity and principal angle similarity. If $\mathbf{P}_i$'s are learned by \textbf{Sub}, we denote the whole similarity measure (including exemplars computing and subspace learning) as \textbf{Cos+Sub-PM}. If $\mathbf{P}_i$'s are learned by the proposed \textbf{QSub}, we denote the similarity as \textbf{Cos+QSub-PM}.

\subsubsection{Quality-Aware Exemplars}
In either \textbf{Cos+Sub-PM} or \textbf{Cos+QSub-PM} we are still using simple average pooling to compute the exemplars. But as discussed in Section~\ref{Subspace}, templates consist of faces of different quality. Treating them equally for pooling will let some low-quality faces degrade the global representation of the template. Therefore, we propose to use the same normalized detection score as in Section~\ref{Subspace} to compute the quality-aware exemplars by $\mathbf{e}_D=\frac{1}{L}\sum_{i=1}^L\tilde{d}_i\mathbf{y}_{i}$,
where $\tilde{d}_i=softmax(ql_i)$ and $l_i$ are computed by \eqref{li}.
Then, the cosine distance between the quality-aware exemplars is
\begin{equation}\label{cosd}
s_{QCos}(\mathbf{Y}_1,\mathbf{Y}_2)=\frac{\mathbf{e}_{D1}^T\mathbf{e}_{D2}}{\|\mathbf{e}_{D1}\|_2\|\mathbf{e}_{D2}\|_2}
\end{equation}
and we denote it as \textbf{QCos}. Using the new cosine distance, the similarity becomes
\begin{align}
s(\mathbf{Y}_1,\mathbf{Y}_2)&=s_{QCos}(\mathbf{Y}_1,\mathbf{Y}_2)+\lambda s_{PM}(\mathbf{P}_1, \mathbf{P}_2)
\end{align}
If $P_i$'s are learned by \textbf{QSub}, the similarity is further denoted by \textbf{QCos+QSub-PM}.
\subsubsection{Variance-Aware Projection Metric}
As previously discussed, the projection metric $S_{PM}(S_1, S_2)$ is the square root of the mean square of principle angles between two subspaces and it treats each basis in each subspace equally. But these bases are actually eigenvectors of an eigenvalue decomposition problem. Different basis corresponds to different eigenvalue, which represents the variance of data in each basis direction. Obviously, those bases with larger variances contain more information than those with smaller variances. Therefore, based on the variance of each basis, we propose a variance-aware projection metric as:
\begin{equation}\label{VPM}
s_{VPM}(\mathbf{P}_1,\mathbf{P}_2)=\sqrt{\frac{1}{r}\|\tilde{\mathbf{P}}_1^T\tilde{\mathbf{P}}_2\|_F^2}
\end{equation}
where
\begin{equation}
\tilde{\mathbf{P}}_i = \frac{1}{tr(\log(\boldsymbol{\Lambda}_i))}\mathbf{P}_i\log(\boldsymbol{\Lambda}_i)
\end{equation}
$\boldsymbol{\Lambda}_i$ is a diagonal matrix whose diagonals are eigenvalues corresponding to eigenvectors in $\mathbf{P}_i$. $\frac{1}{tr(\log(\boldsymbol{\Lambda}_i))}$ is the normalization factor. We use the logarithm of variance to weight different bases in a subspace. This similarity measure is inspired by the Log-Euclidean distance used for image-set classification in \cite{cov}. Empirically, we use $\max(0, \log(\boldsymbol{\Lambda}_i))$ instead of $\log(\boldsymbol{\Lambda}_i)$ to avoid negative weights. We use \textbf{VPM} to denote this similarity measure.

\subsubsection{Quality-Aware Subspace-to-Subspace Similarity}
By combining the quality-aware subspace learning, quality-aware exemplars and variance-aware projection metric, we propose the quality-aware subspace-to-subspace similarity between two video templates as:
\begin{align}
s(\mathbf{Y}_1,\mathbf{Y}_2)&=s_{QCos}(\mathbf{Y}_1,\mathbf{Y}_2) + \lambda s_{VPM}(\mathbf{P}_{D1},\mathbf{P}_{D2})
\end{align}
where $s_{QCos}$ is defined in \eqref{cosd}, $\mathbf{P}_{Di}$'s are learned by \eqref{PD} and $s_{VPM}$ is defined in \eqref{VPM}. This similarity measure is denoted as \textbf{QCos+QSub-VPM}. 
Comparisons of the proposed similarity measures and other baselines on several challenging datasets will be discussed in Section~\ref{sec:exp}. 
\section{Experiments}\label{sec:exp}
In this section, we report video-based face recognition results for the proposed system on two challenging unconstrained multimedia face recognition dataset, IARPA Janus Benchmark B (IJB-B) \cite{ijbb} and IARPA Janus Surveillance Video Benchmark  (IJB-S), and compare with other baseline methods. We also provide results on Multiple Biometric Grand Challenge (MBGC) Version 1 \cite{MBGC}, and Face and Ocular Challenge Series (FOCS) \cite{FOCS} datasets, to demonstrate the effectiveness of the proposed system. We discuss below the details of datasets, protocols and our training and testing procedures.

\subsection{Datasets}
\textbf{IARPA Janus Benchmark B (IJB-B):} IJB-B \cite{ijbb} dataset is an unconstrained face recognition dataset. It contains 1845 subjects with 11,754 images, 55,025 frames and 7,011 multiple-shot videos. IJB-B is a template-based dataset where a template consists of a varying number of still images or video frames from different sources. A template can be either image-only, or video-frame-only, or mixed media template.  Sample frames from this dataset are shown in Figure~\ref{fig:ijbb_example}.

In this paper, we only focus on the 1:N video protocol of IJB-B. It is an open set 1:N identification protocol where each given probe is collected from a video and is searched among all gallery faces. Gallery candidates are ranked according to their similarity scores to the probes. Top-K rank accuracy and True Positive Identification Rate (TPIR) over False Positive Identification Rate(FPIR) are used to evaluate the performance. The gallery templates are separated into two splits, $G_1$ and $G_2$,  all consisting of still images. For each video, we are given the frame index with face bounding box of the first occurrence of the target subject, as shown in Figure~\ref{fig:ijbb_example}. Based on this anchor, all the faces in that video with the same identity should be collected to construct the probes. The identity of the first occurrence bounding box will be considered as the template identity for evaluation.

\textbf{IARPA Janus Surveillance Video Benchmark (IJB-S):} Similar to IJB-B, the IJB-S dataset is also a template-based, unconstrained video face recognition dataset. It contains faces in two separate domains: high-resolution still images for galleries and low quality, remotely captured surveillance videos for probes. It consists of 202 subjects from 1421 images and 398 single-shot surveillance videos. The number of subjects is small compared to IJB-B, but it is even more challenging due to the low quality nature of surveillance videos.

Based on the choices of galleries and probes, we are interested in three different surveillance video-based face recognition protocols: surveillance-to-single protocol, surveillance-to-booking protocol and surveillance-to-surveillance protocol. These are all open set 1:N protocols where each probe is searched among the given galleries. Like IJB-B, the probe templates are collected from videos, but no annotations are provided. Thus raw face detections should be grouped to construct templates with the same identities.

Galleries consist of only single frontal high resolution image for surveillance-to-single protocol. Galleries are constructed by both frontal and multiple-pose high resolution images for surveillance-to-booking protocol. For the most challenging surveillance-to-surveillance protocol, galleries are collected from surveillance videos as well, with given bounding boxes. In all three protocols, gallery templates are split into two splits, $G_1$ and $G_2$. During evaluation, the detected faces in videos are first matched to the ground truth bounding boxes to find their corresponding identity information. The majority of identities appears in each template will be considered as the identity of the template, and will be used for further identification evaluation. Example frames are shown in Figure~\ref{fig:cs6_example}. Notice the remote faces are acquired with very low quality.

\begin{figure}[t]
\centering
\begin{subfigure}[b]{0.23\textwidth}
\centering
\includegraphics[width=\linewidth]{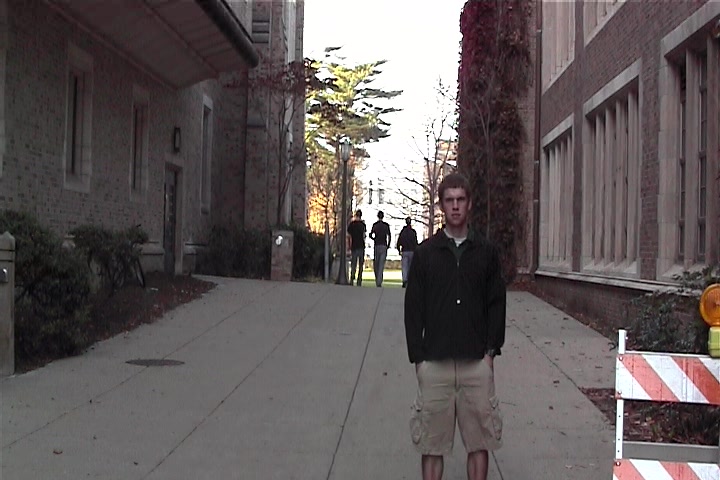}
\caption{MBGC Walking}
\end{subfigure}
~
\begin{subfigure}[b]{0.23\textwidth}
\centering
\includegraphics[width=\linewidth]{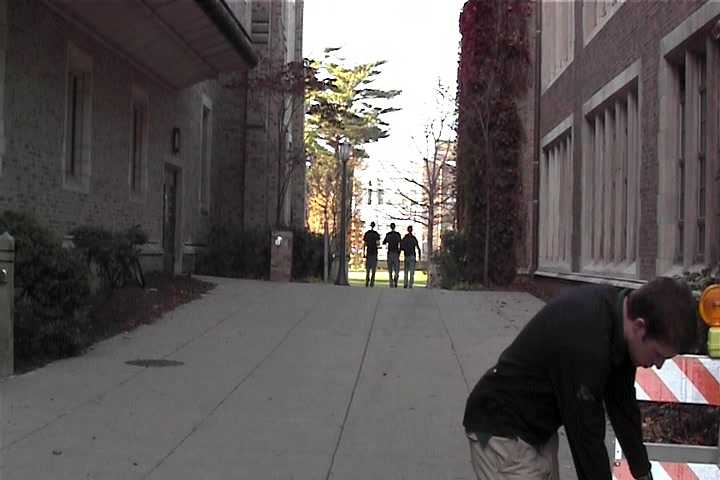}
\caption{MBGC Activity}
\end{subfigure}
\\
\begin{subfigure}[b]{0.23\textwidth}
\centering
\includegraphics[width=\linewidth]{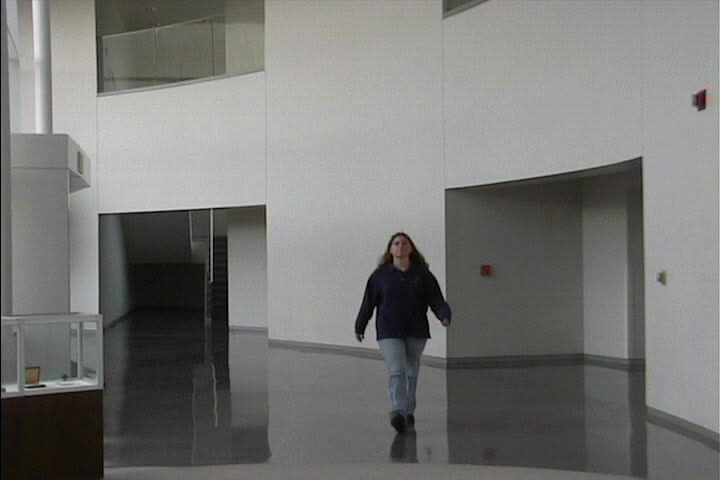}
\caption{FOCS Walking}
\end{subfigure}
~
\begin{subfigure}[b]{0.23\textwidth}
\centering
\includegraphics[width=\linewidth]{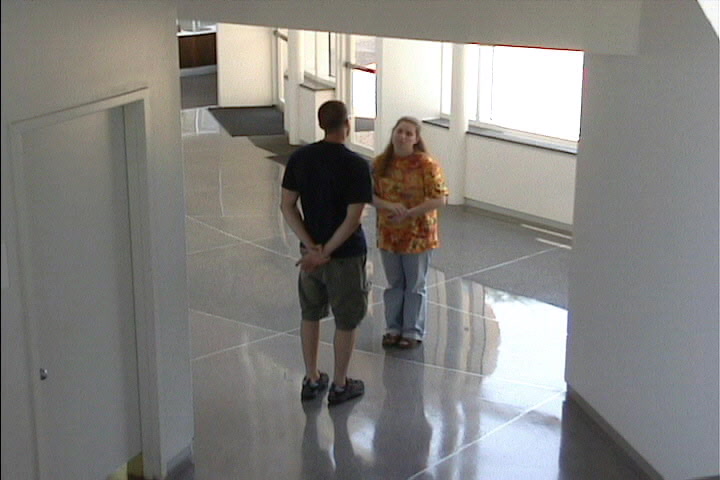}
\caption{FOCS Activity}
\end{subfigure}
\caption{Examples of MBGC and FOCS datasets}
\label{fig:mbgc_focs_example}
\end{figure}

\textbf{Multiple Biometric Grand Challenge (MBGC):} The MBGC Version 1 dataset contains 399 walking (frontal face) and 371 activity (profile face) video sequences from 146 people. Figure~\ref{fig:mbgc_focs_example} shows some sample frames from different walking and activity videos. In the testing protocol, verification is specified by two sets: target and query. The protocol requires the algorithm to match each target sequence with all query sequences. Three verification experiments are defined: walking-vs-walking (WW), activity-vs-activity (AA) and activity-vs-walking (AW). 

\textbf{Face and Ocular Challenge Series (FOCS):}
The video challenge of FOCS \cite{FOCS} is designed for frontal and non-frontal video sequence matching. The FOCS UT Dallas dataset contains 510 walking (frontal face) and 506 activity (non-frontal face) video sequences of 295 subjects with frame size of 720$\times$480 pixels. Like MBGC, FOCS specifies three verification protocols: walking-vs-walking, activity-vs-walking, and activity-vs-activity. In these experiments, 481 walking videos and 477 activity videos are chosen as query videos. The size of target sets ranges from 109 to 135 video sequences. Sample video frames from this dataset are shown in Figure~\ref{fig:mbgc_focs_example}.

\begin{figure*}[t!]
\centering
\begin{subfigure}[b]{0.32\textwidth}
\includegraphics[width=\textwidth]{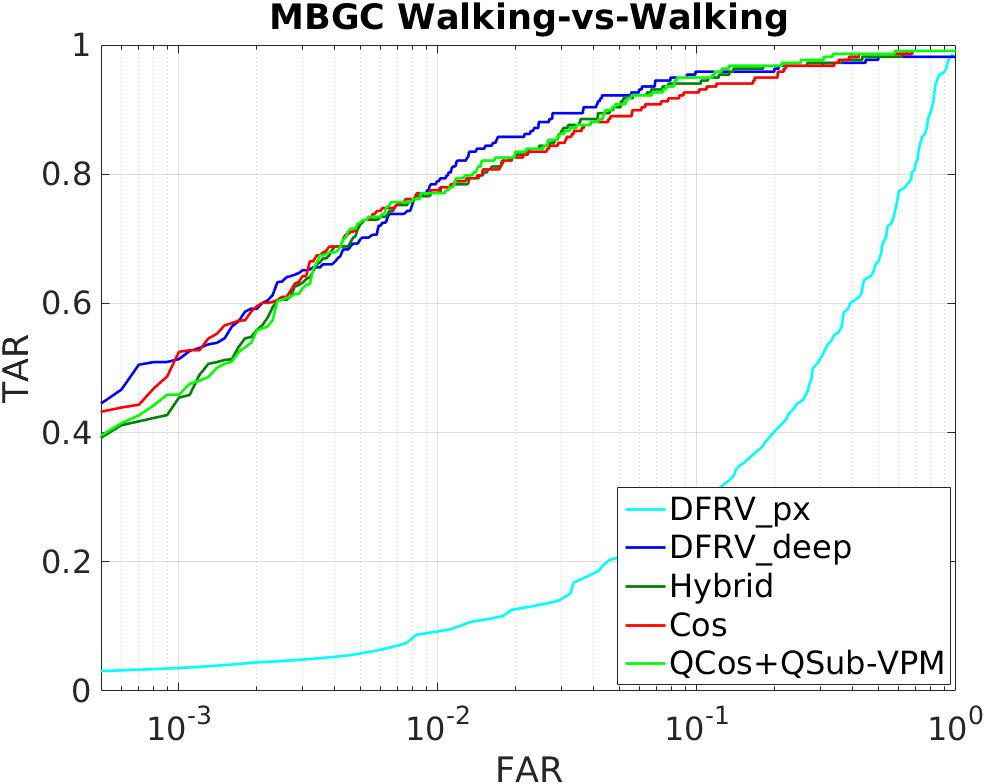}
\end{subfigure}
~
\begin{subfigure}[b]{0.32\textwidth}
\includegraphics[width=\textwidth]{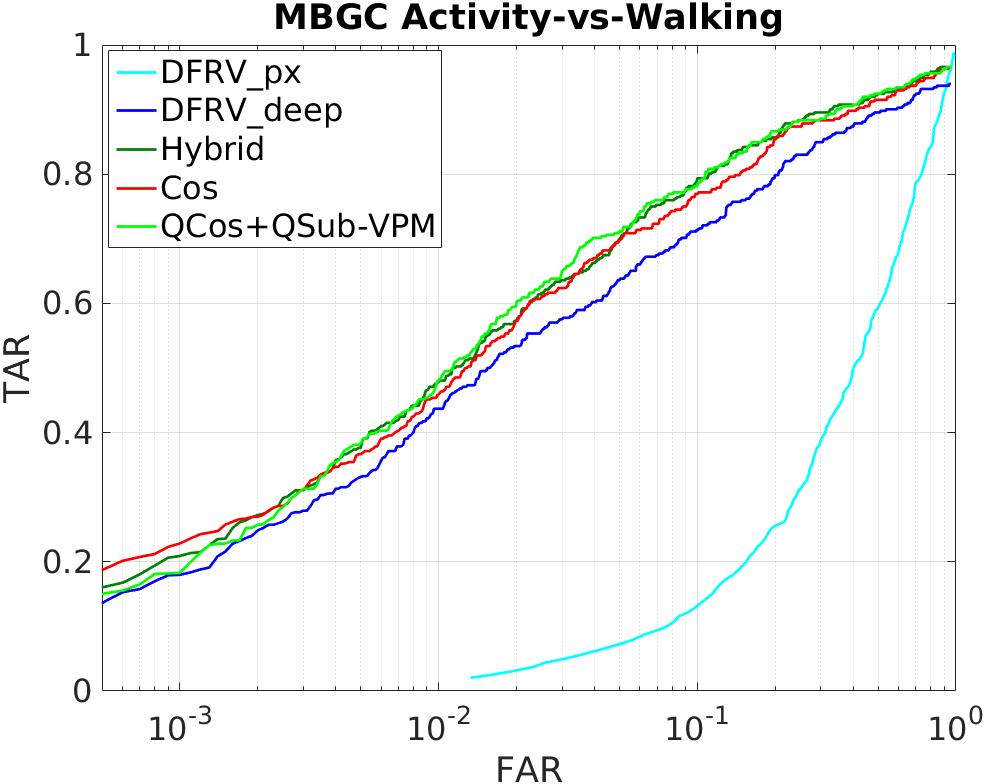}
\end{subfigure}
~
\begin{subfigure}[b]{0.32\textwidth}
\includegraphics[width=\textwidth]{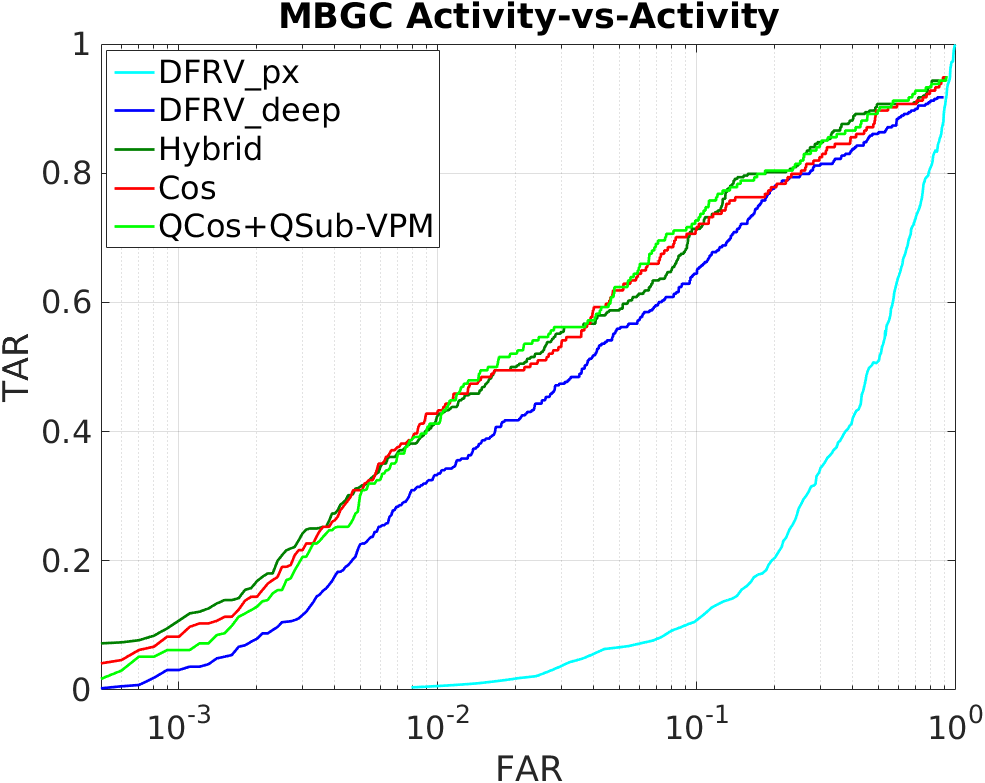}
\end{subfigure}
\\
\begin{subfigure}[b]{0.32\textwidth}
\includegraphics[width=\textwidth]{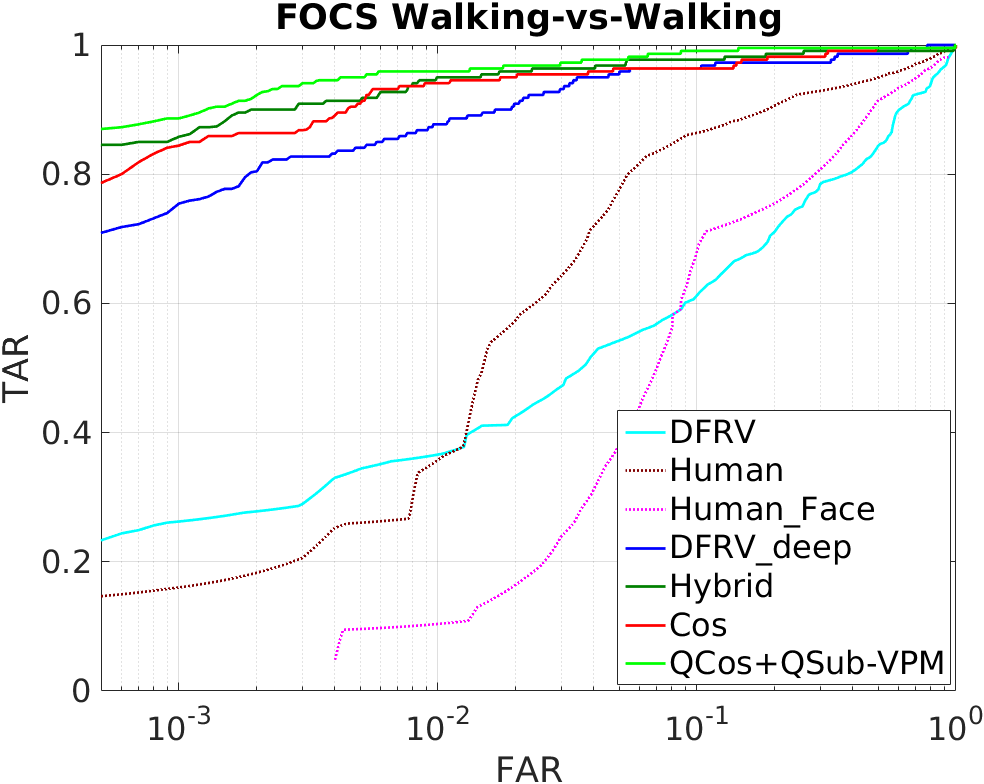}
\end{subfigure}
~
\begin{subfigure}[b]{0.32\textwidth}
\includegraphics[width=\textwidth]{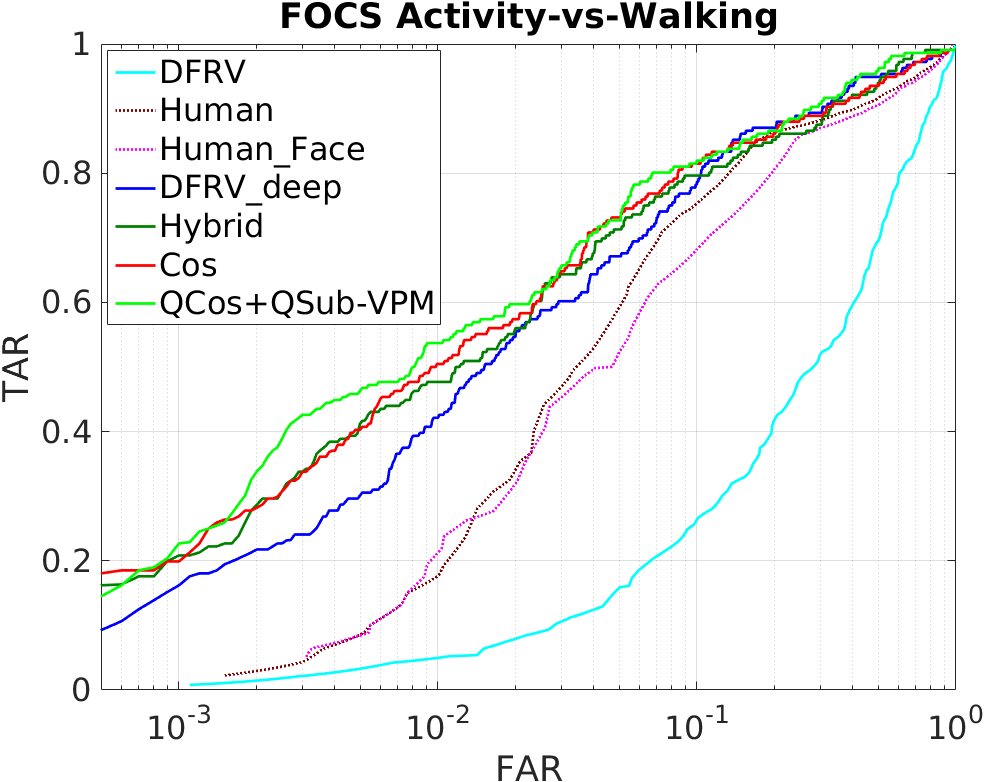}
\end{subfigure}
~
\begin{subfigure}[b]{0.32\textwidth}
\includegraphics[width=\textwidth]{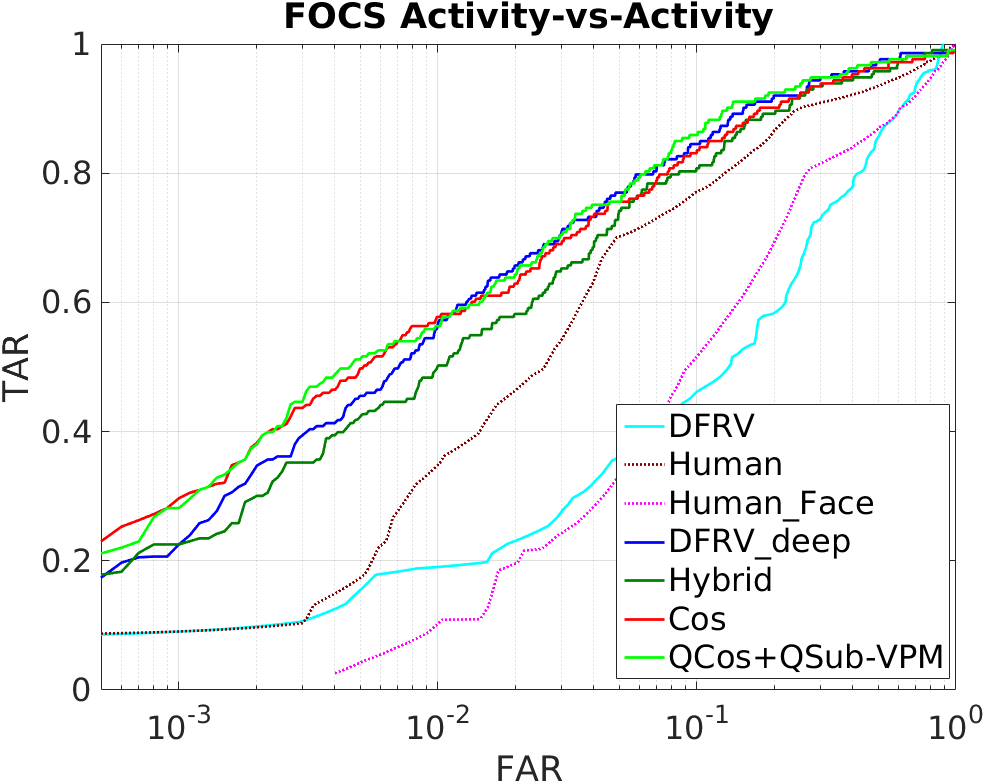}
\end{subfigure}
\caption{Verification results on MBGC and FOCS datasets}
\label{fig:mbgc_focs_v}
\end{figure*}

\subsection{Implementation Details}\label{sec:imp}
In this part, we discuss the implementation details for each dataset respectively.
\subsubsection{IJB-B}
For the IJB-B dataset, we employ the SSD face detector \cite{icipssd} to extract the face bounding boxes in all images and video frames. We employ the facial landmark branch of All-in-One Face \cite{allinone} for fiducial detection on every detected bounding boxes and apply facial alignment based on these fiducials using the seven-point similarity transform.

The aligned faces are further represented using three networks proposed in \cite{magazine}. We denote them as Network A, Network B and Network C. Network A modifies the ResNet-101 \cite{He2015residual} architecture. It has an input size of dimensions $224\times224$ and adds an extra fully connected layer after the last convolutional layer to reduce the feature dimensionality to 512. Also it replaces the original softmax loss with the crystal loss \cite{crystal} for more stable training. Network B uses the Inception-ResNet-v2 \cite{inception} model as the base network. Similar to Network A, an additional fully-connected layer is added for dimensionality reduction. Naive softmax followed by cross entropy loss is used for this network. Network C is based on the face recognition branch in the All-in-One Face architecture \cite{allinone}. The branch consists of seven convolutional layers followed by three fully connected layers.

Network A and Network C are trained on the MSCeleb-1M dataset \cite{msceleb} which contains 3.7 million images from 57,440 subjects. Network B is trained on the union of three datasets called the Universe dataset: 3.7 million still images from the MSCeleb-1M dataset, 300,000 still images from the UMDFaces dataset \cite{UMDFaces}, and about 1.8 million video frames from the UMDFaces Video dataset. For each network, we further reduce its dimensionality into 128 by triplet probabilistic embedding (TPE) \cite{SwamiBTAS} trained on the UMDFaces dataset.

For face association, we use the kernelized correlation filter (KCF) tracker \cite{ofstracker} to pre-associate the initial face bounding boxes for target subjects. The tracking algorithm is applied on the face detection bounding boxes for the first k = 50 frames after the annotated frame. If a pre-associated face detection bounding box has an IoU with the tracking bounding box less than 0.3, we discard the bounding box to prevent undesirable pre-associations resulting from the drifting of the tracker. For the parameters of face association, two bounding boxes with IoU ratio less than $\gamma = 0.1$ are enforced by a cannot-link constraint.

A dataset of 160,498 face images from 1,710 subjects is collected to model the negative background subjects. 
We use the weighted LIBLINEAR implementation \cite{liblinear} with L2-regularized L2-loss support vector classification setting to learn the weight vector, and its cost parameter C is set to 10.

Then, features from associated bounding boxes are used to construct the probe templates. We use quality-aware pooling for both gallery and probe templates to calculate their exemplars (\textbf{QCos}) where $t = 7$ and $q=0.3$ are used for detection score normalization. Subspaces are built by applying quality-aware subspace learning method (\textbf{QSub}) on each template and taking the top three eigenvector with the largest corresponding eigenvalues. When fusing the cosine similarity and variance-aware projection similarity metric (\textbf{VPM}), we use $\lambda = 1$ so two similarity scores are fused equally. We compute the subspace-to-subspace similarity score for each network independently, and combine the similarity scores from three networks by score-level fusion. We also implement baseline methods using combinations of exemplars from vanilla average pooling (\textbf{Cos}), subspaces learned by regular PCA (\textbf{Sub}) and projection similarity metric (\textbf{PM}).

\begin{table*}[t]
\centering\resizebox{0.8\textwidth}{!}
{\begin{tabular}{|c|c|c|c|c|c|c|c|c|c|}
 \hline
 Methods &  Rank=1 & Rank=2 & Rank=5 & Rank=10 & Rank=20 & Rank=50 & FPIR=0.1 & FPIR=0.01 \\
 \hline
\cite{TFA} with Iteration 0 & 55.94\% & - & 68.40\% & 72.89\% & - &  83.71\% & 44.60\% & 28.73\%\\
 \hline
\cite{TFA} with Iteration 3 & 61.01\% & - & 73.39\% & 77.90\% & - &  87.62\% & 49.73\% & 34.11\%\\
 \hline
 \cite{TFA} with Iteration 5 & 61.00\% & - & 73.46\% & 77.94\% & - &  87.69\% & 49.78\% & 33.93\%\\
 \hline
 \hline
Cos & 78.37\% & 81.35\% & 84.39\% & 86.29\% & 88.30\% & 90.82\% & 73.15\% & 52.19\%\\
\hline
QCos & 78.43\% & 81.41\% & 84.40\% & 86.33\% & 88.34\% & 90.88\% & \textbf{73.19\%} & \textbf{52.47\%}\\
\hline
Cos+Sub-PM & 77.99\% & 81.45\% & 84.68\% & 86.75\% & 88.96\% & 91.91\% & 72.31\% & 38.44\%\\
\hline
QCos+Sub-PM & 78.02\% & 81.46\% & 84.76\% & 86.72\% & 88.97\% & 91.91\% & 72.38\% & 38.88\%\\
\hline
QCos+QSub-PM & 78.04\% & 81.47\% & 84.73\% & 86.72\% & 88.97\% & 91.93\% & 72.39\% & 38.91\%\\
\hline
QCos+QSub-VPM & \textbf{78.93\%} & \textbf{81.99\%} & \textbf{84.96\%} & \textbf{87.03\%} & \textbf{89.24\%} & \textbf{92.02\%} & 71.26\% & 47.35\%\\
\hline
\end{tabular}}
\caption{1:N Search Top-K Average Accuracy and TPIR/FPIR of IJB-B video search protocol}
\label{tab:ijbb}
\end{table*}

\subsubsection{IJB-S}
For the IJB-S dataset, we employ the multi-scale face detector DPSSD to detect faces in surveillance videos. We only keep face bounding boxes with detection scores greater than 0.4771, to reduce the number of false detections. We use the facial landmark branch of All-in-One Face \cite{allinone} as the fiducial detector. Face alignment is performed using the seven-point similarity transform.

Different from IJB-B, since IJB-S does not specify the subject of interest, we are required to localize and associate all the faces for different subjects to yield the probe sets. Since IJB-S videos are single-shot, we use SORT \cite{SORT} to track every face appearing in the videos. Faces in the same tracklet are grouped to create a probe template. Since some faces in surveillance videos are of extreme pose, blur and low-resolution, to improve precision, tracklets consisting of such faces should be rejected during the recognition stage. By observation, we find that most of the short tracklets are of low quality and not reliable. The average of the detection score provided by DPSSD is also used as an indicator of the quality of the tracklet. On the other hand, we also want to take the performance of face detection into consideration to strike a balance between recall and precision. Thus in our experiments, we use two configurations for tracklets filtering: 1) We keep those tracklets with length greater than or equal to 25 and average detection score greater than or equal to 0.9 to reject low-quality tracklets and focusing on precision. It is referred to as \textbf{with Filtering}. 2) Following the settings in \cite{ijbs}, we produce results without any tracklets filtering and focusing on both precision and recall. It is referred to as \textbf{without Filtering}.

Because of the remote acquisition scenario and blurred probes in the IJB-S dataset, we retrain Network A with the same crystal loss but on the Universe dataset used by Network B. We denote it as Network D. We also retrain Network B with the crystal loss \cite{crystal} on the same training data. We denote it as Network E. As a combination of high capacity network and large scale training data, Network D and E are more powerful than Networks A, B, and C. As before, we reduce feature dimensionality into 128 using the TPE trained on the UMDFaces dataset.

In IJB-S, feature aggregation and matching parts are the same as IJB-B except that we combine the similarity score  by score-level fusion from Network D and E. One thing that needs to be mentioned is that for the surveillance-to-surveillance protocol, we only use single Network D for representation as Network E is ineffective for low-quality gallery faces in this protocol.

\subsubsection{MBGC and FOCS}
For MBGC and FOCS datasets, we use All-in-One Face for both face detection and facial landmark detection. The MBGC and FOCS datasets contain only one person in a video in general. Hence, for each frame, we directly use the face bounding box with the highest detection score as the target face. Similar to IJB-S, bounding boxes are filtered based on detection scores. From the detected faces, deep features are extracted using Network D. Since MBGC and FOCS datasets do not provide training data, we also use the TPE trained on UMDFaces dataset to reduce feature dimensionality into 128. For MBGC and FOCS, feature aggregation and matching parts are the same as IJB-B and IJB-S.

\subsection{Evaluation Results}
In the following section, we first show some face association results on IJB-B and IJB-S datasets. Then we compare the performance of the proposed face recognition system with several baseline methods. For each dataset, all the baseline methods listed below use deep features extracted from the same network and with the same face detector.
\begin{itemize}
\item \textbf{Cos:} We compute the cosine similarity scores directly from the average pooled exemplars of the deep representation.
\item \textbf{QCos:} We compute the cosine similarity scores from the quality-aware average pooled exemplars of the deep representation.
\item \textbf{Cos+Sub-PM:} Subspace-to-subspace similarity is computed by fusing the plain cosine similarity and plain projection metric, and subspaces are learned by plain PCA.
\item \textbf{QCos+Sub-PM:} Subspace-to-subspace similarity is computed by fusing the quality-aware cosine similarity and plain projection metric, and subspaces are learned by plain PCA.
\item \textbf{QCos+QSub-PM:} Subspace-to-subspace similarity is computed by fusing the quality-aware cosine similarity and plain projection metric, and subspaces are learned by quality-aware subspace learning.
\item \textbf{QCos+QSub-VPM:} Subspace-to-subspace similarity is computed by fusing the quality-aware cosine similarity and variance-aware projection metric, and subspaces are learned by quality-aware subspace learning.
\end{itemize}

\begin{figure}[h]
\centering
\begin{subfigure}[b]{0.95\linewidth}
\centering
\includegraphics[width=\linewidth]{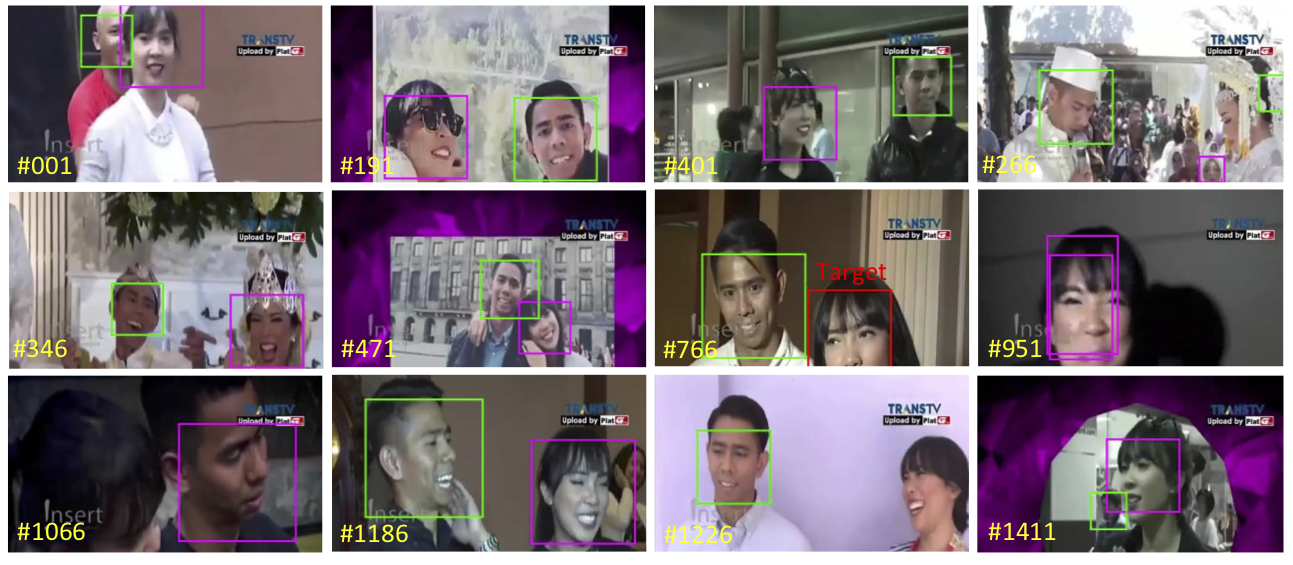}
\end{subfigure}
\\
\begin{subfigure}[b]{0.95\linewidth}
\centering
\includegraphics[width=\linewidth]{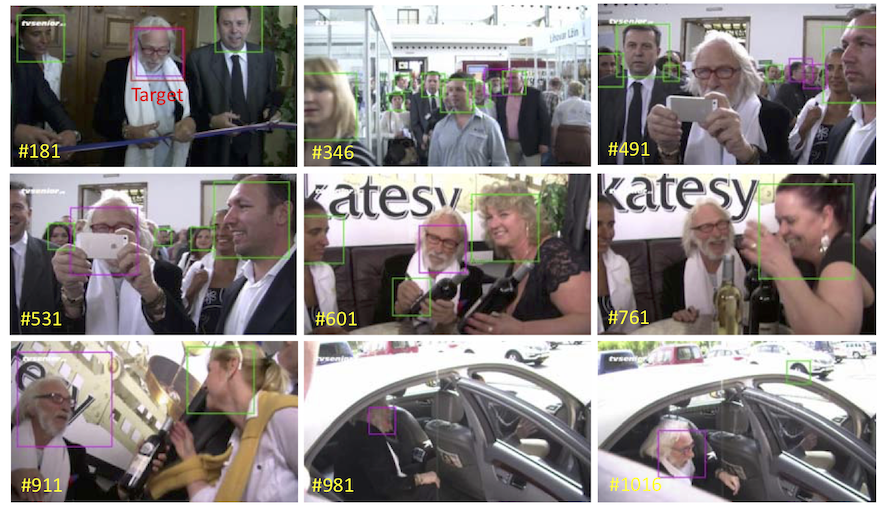}
\end{subfigure}
\caption{Examples of face association results by the proposed TFA method on IJB-B. The target annotation is in red box, and the associated faces of the target subject are in magenta boxes.}
\label{fig:ijbb_result}
\end{figure}

\begin{figure}[h]
\centering
\begin{subfigure}[b]{0.95\linewidth}
\centering
\includegraphics[width=\linewidth]{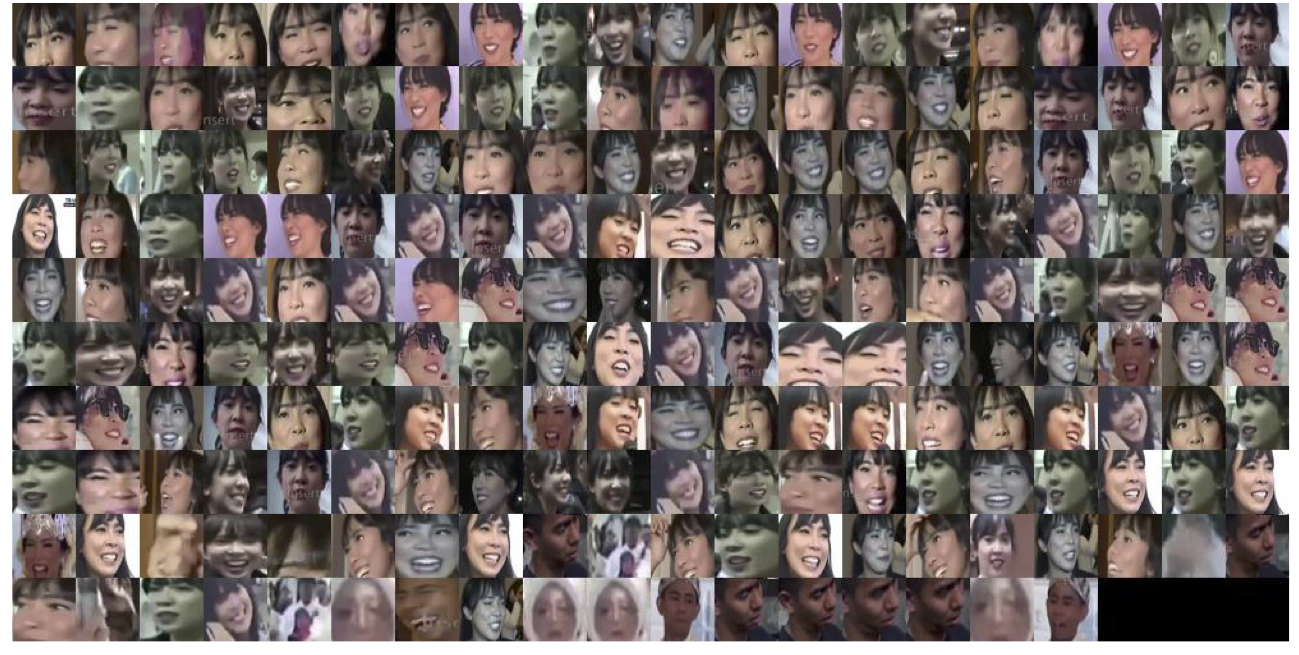}
\end{subfigure}
\\
\begin{subfigure}[b]{0.95\linewidth}
\centering
\includegraphics[width=\linewidth]{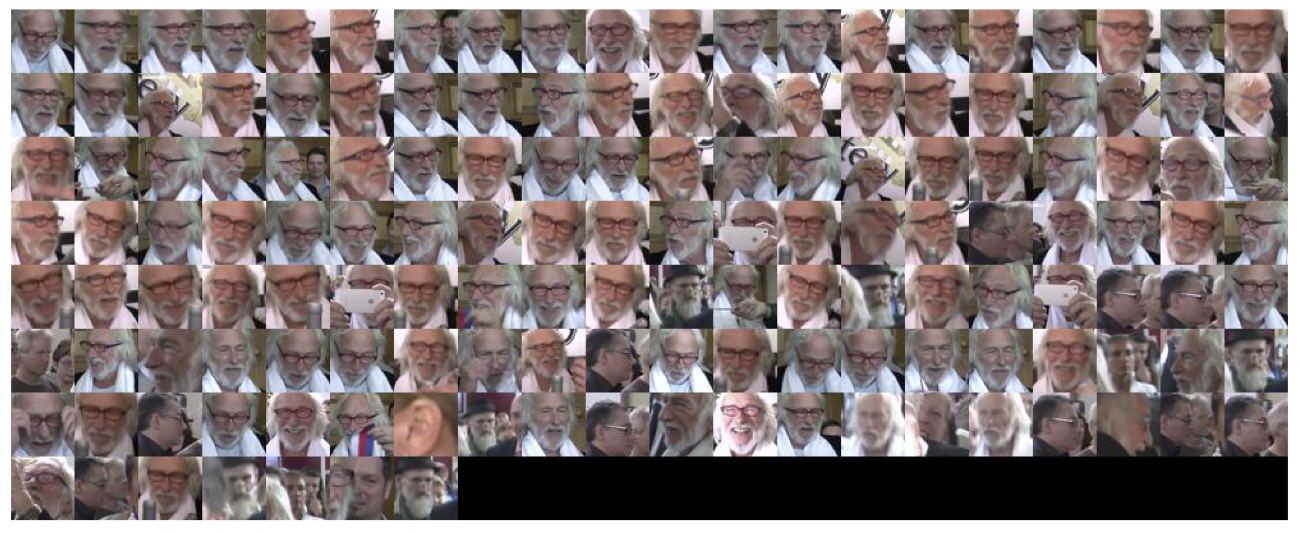}
\end{subfigure}
\caption{Associated faces by the proposed TFA method corresponding to examples in Figure~\ref{fig:ijbb_result}. Face images are in the order of the confidence of face association.}
\label{fig:ijbb_result_box}
\end{figure}

\begin{figure}[h]
\centering
\begin{subfigure}[b]{0.95\linewidth}
\centering
\includegraphics[width=\linewidth]{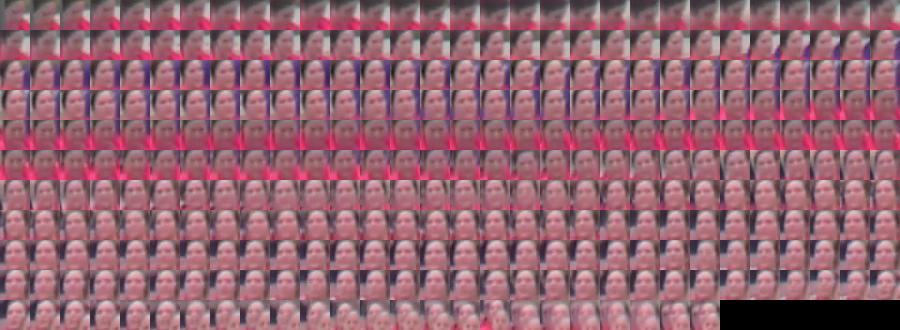}
\end{subfigure}
\\
\vspace{2pt}
\begin{subfigure}[b]{0.95\linewidth}
\centering
\includegraphics[width=\linewidth]{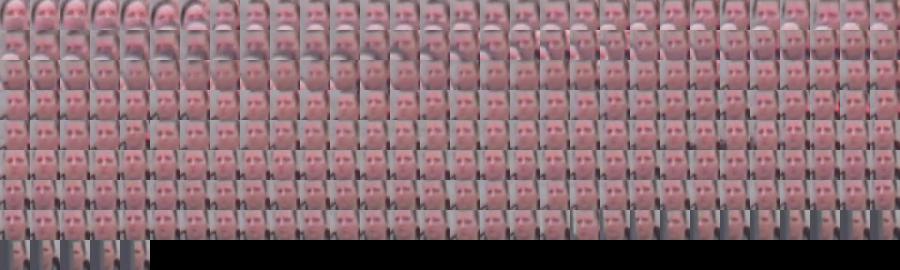}
\end{subfigure}
\caption{Associated faces using SORT in IJB-S. Face images are in their temporal order. Notice the low-quality faces at the boundaries of tracklets since the tracker cannot reliably track anymore.}
\label{fig:cs6_result_box}
\end{figure}

\noindent \textbf{IJB-B:} Figures~\ref{fig:ijbb_result} and \ref{fig:ijbb_result_box} show some examples of our face association results using the proposed TFA in IJB-B dataset. Table~\ref{tab:ijbb} shows the Top-K Accuracy results for IJB-B video protocol. In this dataset, besides the baselines, our method is compared with original results in \cite{TFA} corresponding to different iteration numbers. Results shown are the average of two galleries. Notice that our proposed system and \cite{TFA} use the same face association method, but we have different networks and feature representation techniques.

\begin{table*}[t]
\centering\resizebox{\textwidth}{!}
{\begin{tabular}{|c|c|c|c|c|c|c|c|c|c|c|c|c|}
 \hline
  \multirow{2}{*}{Methods} & \multicolumn{6}{c|}{Top-K Average Accuracy \textbf{with Filtering}} & \multicolumn{6}{c|}{EERR metric \textbf{without Filtering}}\\ 
 \cline{2-13}
 & R=1 & R=2 & R=5 & R=10 & R=20 & R=50 & R=1 & R=2 & R=5 & R=10 & R=20 & R=50\\
 \hline
Arc-Cos \cite{arcface} & 52.03\% & 56.83\% & 63.16\% & 69.05\% & 76.13\% & 88.95\% & 24.45\% & 26.54\% & 29.35\% & 32.33\% & 36.38\% & 44.81\%\\
\hline
Arc-QCos+QSub-PM & 60.92\% & 65.06\% & 70.45\% & 75.19\% & 80.69\% & 90.29\% & 28.73\% & 30.44\% & 32.98\% & 35.40\% & 38.70\% & 45.46\%\\
 \hline
 \hline
Cos & 64.86\% & 70.87\% & 77.09\% & 81.53\% & 86.11\% & 93.24\% & 29.62\% & 32.34\% & 35.60\% & 38.36\% & 41.53\% & 46.78\%\\
\hline
QCos & 65.42\% & 71.34\% & 77.37\% & 81.78\% & 86.25\% & 93.29\% & 29.94\% & 32.60\% & 35.85\% & 38.52\% & 41.70\% & 46.78\%\\
\hline
Cos+Sub-PM & 69.52\% & 75.15\% & 80.41\% & 84.14\% & 87.83\% & 94.27\% & 32.22\% & 34.70\% & 37.66\% & 39.91\% & 42.65\% & 47.54\%\\
\hline
QCos+Sub-PM & 69.65\% & 75.26\% & 80.43\% & 84.22\% & 87.81\% & 94.25\% & 32.27\% & 34.73\% & 37.66\% & 39.91\% & 42.67\% & 47.54\%\\
\hline
QCos+QSub-PM & \textbf{69.82\%} & \textbf{75.38\%} & \textbf{80.54\%} & \textbf{84.36}\% & \textbf{87.91\%} & \textbf{94.34\%} & \textbf{32.43\%} & \textbf{34.89\%} & \textbf{37.74\%} & \textbf{40.01\%} & \textbf{42.77\%} & \textbf{47.60\%}\\
\hline
QCos+QSub-VPM & 69.43\% & 75.24\% & 80.34\% & 84.14\% & 87.86\% & 94.28\% & 32.19\% & 34.75\% & 37.68\% & 39.88\% & 42.56\% & 47.50\%\\
\hline
\end{tabular}}
\caption{1:N Search results of IJB-S surveillance-to-single protocol. Using both Network D and E for representation.}
\label{tab:cs6_sg2s}
\end{table*}

\begin{table*}[t]
\centering\resizebox{\textwidth}{!}
{\begin{tabular}{|c|c|c|c|c|c|c|c|c|c|c|c|c|}
 \hline
  \multirow{2}{*}{Methods} & \multicolumn{6}{c|}{Top-K Average Accuracy \textbf{with Filtering}} & \multicolumn{6}{c|}{EERR metric \textbf{without Filtering}}\\ 
 \cline{2-13}
 & R=1 & R=2 & R=5 & R=10 & R=20 & R=50 & R=1 & R=2 & R=5 & R=10 & R=20 & R=50\\
 \hline
Arc-Cos \cite{arcface} & 54.59\% & 59.12\% & 65.43\% & 71.05\% & 77.84\% & 89.16\% & 25.38\% & 27.58\% & 30.59\% & 33.42\% & 37.60\% & 45.05\%\\
\hline
Arc-QCos+QSub-VPM & 60.86\% & 65.36\% & 71.30\% & 76.15\% & 81.63\% & 90.70\% & 28.66\% & 30.64\% & 33.43\% & 36.11\% & 39.57\% & 45.70\%\\
 \hline
 \hline
Cos & 66.48\% & 71.98\% & 77.80\% & 82.25\% & 86.56\% & 93.41\% & 30.38\% & 32.91\% & 36.15\% & 38.77\% & 41.86\% & 46.79\%\\
\hline
QCos & 66.94\% & 72.41\% & 78.04\% & 82.37\% & 86.63\% & 93.43\% & 30.66\% & 33.17\% & 36.28\% & 38.84\% & 41.88\% & 46.84\%\\
\hline
Cos+Sub-PM & 69.39\% & 74.55\% & 80.06\% & 83.91\% & 87.87\% & \textbf{94.34\%} & 32.02\% & 34.42\% & 37.59\% & 39.97\% & 42.64\% & \textbf{47.58\%}\\
\hline
QCos+Sub-PM & 69.57\% & 74.78\% & 80.06\% & 83.89\% & 87.94\% & 94.33\% & 32.16\% & 34.61\% & 37.62\% & 39.99\% & 42.71\% & 47.57\%\\
\hline
QCos+QSub-PM & 69.67\% & 74.85\% & 80.25\% & 84.10\% & 88.04\% & 94.22\% & 32.28\% & 34.77\% & 37.76\% & 40.11\% & \textbf{42.76\%} & 47.57\%\\
\hline
QCos+QSub-VPM & \textbf{69.86\%} & \textbf{75.07\%} & \textbf{80.36\%} & \textbf{84.32\%} & \textbf{88.07\%} & 94.33\% & \textbf{32.44\% }& \textbf{34.93\%} & \textbf{37.80\%} & \textbf{40.14\%} & 42.72\% & \textbf{47.58\%}\\
\hline
\end{tabular}}
\caption{1:N Search results of IJB-S surveillance-to-booking protocol. Using both Network D and E for representation.} 
\label{tab:cs6_b2s}
\end{table*}

\begin{table*}[t]
\centering\resizebox{\textwidth}{!}
{\begin{tabular}{|c|c|c|c|c|c|c|c|c|c|c|c|c|}
 \hline
  \multirow{2}{*}{Methods} & \multicolumn{6}{c|}{Top-K Average Accuracy \textbf{with Filtering}} & \multicolumn{6}{c|}{EERR metric \textbf{without Filtering}}\\ 
 \cline{2-13}
 & R=1 & R=2 & R=5 & R=10 & R=20 & R=50 & R=1 & R=2 & R=5 & R=10 & R=20 & R=50\\
 \hline
Arc-Cos \cite{arcface} & 8.68\% & 12.58\% & 18.79\% & 26.66\% & 39.22\%	&68.19\% & 4.98\% & 7.17\% & 10.86\% & 15.42\% & 22.34\% & 37.68\%\\
\hline
Arc-QCos+QSub-PM & 8.64\% & 12.57\% & 18.84\% & 26.86\% & 39.78\% & \textbf{68.21\%} & \textbf{5.26\%} & \textbf{7.44\%} & \textbf{11.31\%} & 15.90\% & \textbf{22.68\%} & \textbf{37.83\%}\\
 \hline
 \hline
Cos(D+E) & 9.24\% & 12.51\% & 19.36\% & 25.99\% & 32.95\% & 52.95\% & 4.74\% & 6.62\% & 10.70\% & 14.88\% & 19.29\% & 30.64\%\\
 \hline
QCos+QSub-VPM(D+E) & \textbf{9.56\%} & \textbf{13.03\%} & 19.65\% & 27.15\% & 35.39\% & 56.02\% & 4.77\% & 6.78\% & 10.88\% & 15.52\% & 20.51\% & 32.16\%\\
 \hline
 \hline
Cos(D) & 8.54\% & 11.99\% & 19.60\% & 28.00\% & 37.71\% & 59.44\% & 4.42\% & 6.15\% & 10.84\% & 15.73\% & 21.14\% & 33.21\%\\
\hline
QCos(D) & 8.62\% & 12.11\% & 19.62\% & 28.14\% & 37.78\% & 59.21\% & 4.46\% & 6.20\% & 10.80\% & 15.81\% & 21.06\% & 33.17\%\\
\hline
Cos+Sub-PM(D) & 8.19\% & 11.79\% & 19.56\% & 28.62\% & 39.77\% & 63.15\% & 4.26\% & 6.25\% & 10.79\% & 16.18\% & 22.48\% & 34.82\%\\
\hline
QCos+Sub-PM(D) & 8.24\% & 11.82\% & 19.68\% & 28.68\% & 39.68\% & 62.96\% & 4.27\% & 6.25\% & 10.92\% & 16.18\% & 22.39\% & 34.69\%\\
\hline
QCos+QSub-PM(D) & 8.33\% & 11.88\% & 19.82\% & 28.65\% & 39.78\% & 62.79\% & 4.33\% & 6.21\% & 10.96\% & 16.19\% & 22.48\% & 34.69\%\\
\hline
QCos+QSub-VPM(D) & 8.66\% & 12.27\% & \textbf{19.91\%} & \textbf{29.03\%} & \textbf{40.20\%} & 63.20\% & 4.30\% & 6.30\% & 10.99\% & \textbf{16.23\%} & 22.50\% & 34.76\%\\
\hline
\end{tabular}}
\caption{1:N Search results of IJB-S surveillance-to-surveillance protocol. D stands for only using Network D for representation. D+E stands for using both Network D and E for representation.}
\label{tab:cs6_s2s}
\end{table*}

\begin{figure}[h]
\centering
\begin{subfigure}[b]{0.8\linewidth}
\centering
\includegraphics[width=\linewidth]{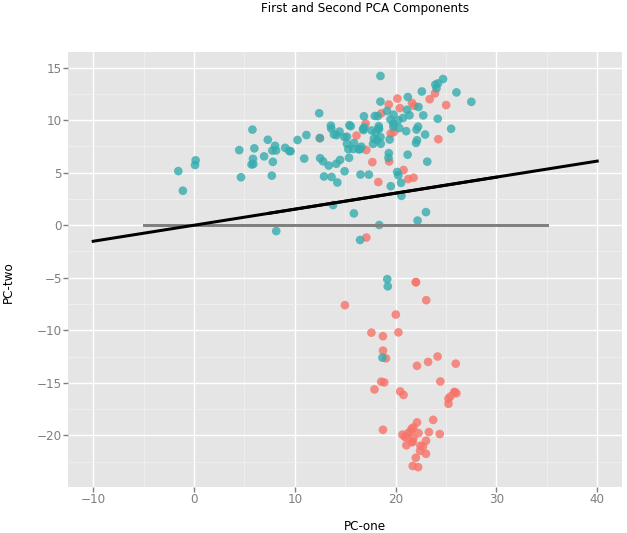}
\end{subfigure}
\begin{subfigure}[b]{0.8\linewidth}
\centering
\includegraphics[width=\linewidth]{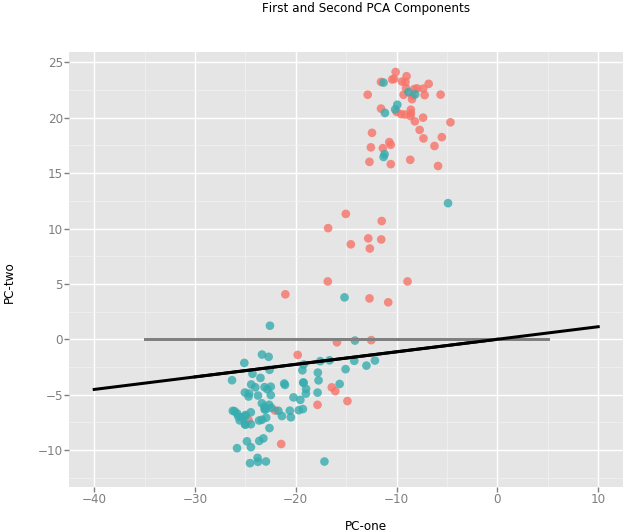}
\end{subfigure}
\caption{Visualization of example templates in IJB-S. Each sample is a dot in the plot with their first two principal components as the coordinates. Samples with $d_i \geq 0.7$ are in \textbf{blue} dots and the rest samples are in \textbf{red} dots. \textbf{Grey} line and \textbf{black} line are the projection of the first subspace basis learned by \textbf{Sub} and \textbf{QSub} respectively.}
\label{fig:cs6_subspace}
\end{figure}

\noindent \textbf{IJB-S:} Figure~\ref{fig:cs6_result_box} shows some examples of our face association results using SORT in IJB-S dataset. Table~\ref{tab:cs6_sg2s}, Table~\ref{tab:cs6_b2s} and Table~\ref{tab:cs6_s2s} show the results for IJB-S surveillance-to-single protocol, surveillance-to-booking protocol and surveillance-to-surveillance protocol respectively. Notice that under the \textbf{with Filtering} configuration, we use the regular top-K average accuracy for evaluation. Under the \textbf{without Filtering} configuration, we use the End-to-End Retrieval Rate (EERR) metric proposed in \cite{ijbs} for evaluation. For surveillance-to-surveillance protocol, we show results for two different network configurations as well. We also implement state-of-the-art network ArcFace \cite{arcface} on IJB-S and compare our method with it. Results from ArcFace are shown with the prefix \textbf{Arc-}. 

Two recent works \cite{cfan} and \cite{rean} have reported results on the IJB-S dataset. These works mainly focused on face recognition and not detection so that they built video templates by matching their detections with ground truth bounding boxes provided by the protocols and evaluated their methods using identification accuracy and not EERR metric. Our system focuses on detection, association and recognition. Therefore after detection, we associate faces across the video frames to build templates without utilizing any ground truth information and evaluate our system using both identification accuracy and EERR metric. Since these two template building procedures are so different, a directly comparison is not meaningful.

\begin{table*}[t]
\centering \resizebox{\textwidth}{!}
{\begin{tabular}{|c|c|c|c|c|c|c|c|c|c|c|c|c|}
 \hline
 \multirow{3}{*}{Methods}  & \multicolumn{6}{c|}{MBGC} & \multicolumn{6}{c|}{FOCS}\\
 \cline{2-13}
 & \multicolumn{2}{c|}{WW} & \multicolumn{2}{c|}{AW} & \multicolumn{2}{c|}{AA} & \multicolumn{2}{c|}{WW} & \multicolumn{2}{c|}{AW} & \multicolumn{2}{c|}{AA}\\ 
 \cline{2-13}
 & FAR=0.01 & FAR=0.1 & FAR=0.01 & FAR=0.1 & FAR=0.01 & FAR=0.1 & FAR=0.01 & FAR=0.1 & FAR=0.01 & FAR=0.1 & FAR=0.01 & FAR=0.1\\
 \hline
 Arc-Cos \cite{arcface} & 84.40\% & 92.20\% & 53.88\% & 75.00\% & 32.47\% & 66.49\% & 98.18\% & \textbf{99.09\%} & 48.61\% & 69.44\% & 48.36\% & 78.87\%\\
 \hline
 Arc-QCos+QSub-PM & \textbf{85.32\%} & 92.20\% & \textbf{55.58\%} & 75.00\% & 32.99\% & 64.43\% & \textbf{98.64\%} & \textbf{99.09\%} & 52.31\% & 74.07\% & 50.23\% & 79.81\%\\
\hline
 \hline
 DFRV$_{deep}$ \cite{DFRV_2012} & 78.90\% & \textbf{95.87\%} & 43.69\% & 71.36\% & 33.51\% & 64.95\% & 87.73\% & 96.36\% & 42.13\% & 78.70\% & 56.81\% & 84.51\%\\
 \hline
 Hybrid \cite{hybrid} & 77.06\% & 94.04\% & 48.06\% & \textbf{79.37\%} & 42.53\% & 71.39\%& 95.00\% & 97.73\% & 47.69\% & 79.63\% & 50.23\% & 80.75\%\\
 \hline
 \hline
 Cos & 77.52\% & 92.66\% & 45.87\% & 76.94\% & \textbf{43.30\%} & 71.65\% & 94.09\% & 96.36\% & 50.46\% & 81.48\% & 57.75\% & 83.57\%\\
 \hline
 QCos & 77.52\% & 92.66\% & 47.57\% & 76.94\% & \textbf{43.30\%} & 71.13\% & 95.91\% & \textbf{99.09\%} & \textbf{53.70\%} & 80.09\% & \textbf{58.22\%} & 83.57\%\\
 \hline
 Cos+Sub-PM & 77.98\% & 94.95\% & 47.57\% & 79.13\% & 41.24\% & 72.68\% & 91.82\% & 97.27\% & 49.07\% & \textbf{83.33\%} & 54.93\% & 85.45\%\\
\hline
QCos+Sub-PM & 77.98\% & 94.95\% & 48.30\% & 78.64\% & 41.75\% & \textbf{73.71\%} & 95.91\% & 98.64\% & 52.78\% & 82.87\% & 55.40\% & \textbf{85.92\%}\\
\hline
QCos+QSub-PM & 77.52\% & 94.95\% & 48.54\% & 78.64\% & 41.75\% & 73.20\% & 95.91\% & \textbf{99.09\%} & 52.31\% & 81.02\% & 55.87\% & \textbf{85.92\%}\\
\hline
QCos+QSub-VPM & 77.06\% & 94.95\% & 48.06\% & 78.16\% & 41.24\% & 72.68\% & 95.91\% & \textbf{99.09\%} & \textbf{53.70\%} & 81.94\% & 56.34\% & \textbf{85.92\%}\\
\hline
\end{tabular}}
\caption{Verification results on MBGC and FOCS datasets}
\label{tab:mbgc_focs_v}
\vspace{-15pt}
\end{table*}

\noindent\textbf{MBGC:} The verification results for the MBGC dataset are shown in Table~\ref{tab:mbgc_focs_v} and Figure~\ref{fig:mbgc_focs_v}. We compare our method with the baseline algorithms, \textbf{Hybrid} \cite{hybrid} and \cite{DFRV_2012} using either raw pixels as \textbf{DFRV$_{px}$} (reported in their paper) or deep features as \textbf{DFRV$_{deep}$} (our implementation). We also report the results of the proposed method applied on the ArcFace features with the prefix \textbf{Arc-}. Figure~\ref{fig:mbgc_focs_v} does not include all the baselines, for a clearer view. The result of \cite{DFRV_2012} is not in the table because the authors did not provide exact numbers in their paper.

\noindent \textbf{FOCS:}  The verification results of FOCS dataset are shown in Table~\ref{tab:mbgc_focs_v} and Figure~\ref{fig:mbgc_focs_v}. O'Toole et al. \cite{OToole2011} evaluated the human performance on this dataset. 
In the figures, \textbf{Human} refers to human performance with all bodies of target subjects seen and \textbf{Human\_Face} refers to performance that only faces of the target subjects are seen. Here besides baseline algorithms and \textbf{Hybrid} \cite{hybrid}, we also compare our method with \cite{DFRV_2012} in either raw pixels as \textbf{DFRV$_{px}$} (reported in their paper) or deep features as \textbf{DFRV$_{deep}$} (our implementation). We also report the results using ArcFace features. 
Similarly, the results of \cite{DFRV_2012} and human performance are not in the table since they did not provide exact numbers.

\subsection{Observation and Discussion}

For the IJB-B dataset, we can see that using the TFA approach, our system produces high quality face association results. We do see some incorrect association results in the examples. But since they are of relatively low quality, our deep subspace representation can filter out these samples, and the proposed system performs consistently better than all the results in \cite{TFA} and the baseline \textbf{Cos} on identification accuracy. For open-set metric TPIR/FPIR, the proposed quality-aware cosine similarity achieves better results, but the proposed subspace similarity metric still performs better than \cite{TFA} with a large margin. For IJB-S results, we have similar observations: our system produces reliable tracking results. The low-quality faces on the tracking boundaries are handled by our robust subspace representation as the proposed system with subspace-to-subspace similarity measure performs better than \textbf{Cos} on surveillance-to-single and surveillance-to-booking protocols, by relatively large margin. It also achieves better accuracy than \textbf{Cos} on the surveillance-to-surveillance protocol. We notice that the fusion of Network D and E does not work well on surveillance-to-surveillance protocol, especially at higher rank accuracy. Such observations are consistent under both tracklets filtering configurations and their corresponding metrics: \textbf{with Filtering} with Top-K average accuracy and \textbf{without Filtering} with the EERR metric. The proposed system also outperforms ArcFace with larger margin in surveillance-to-single and surveillance-to-booking protocols of IJB-S. For MBGC and FOCS datasets, from the tables and plots we can see that in general, the proposed approach performs better than \textbf{Cos} baseline, \textbf{DFRV}$_{deep}$, \textbf{DFRV}$_{px}$ and \textbf{Hybrid}.

Figure~\ref{fig:cs6_subspace} shows the visualization of two templates in IJB-S dataset in PCA-subspace, which illustrates the advantage of the proposed subspace learning method. In the plot, each dot corresponds to a sample in the template, where x- and y-axes correspond to the first two principal components of the samples, learned from each template respectively. Relatively high-quality detections with detection score greater than or equal to 0.7 are represented by blue dots. Relatively low-quality detections with detection score less than 0.7 are represented by red dots. The projections of the first subspace bases learned by \textbf{Sub} and the proposed \textbf{QSub} onto the PCA-subspace are grey and black straight lines in the plot, respectively. From the plot we can see that, with quality-aware subspace learning, the subspaces learned by the proposed method put more weights on the high-quality sample. It fits the high-quality samples better than the low-quality ones. But the plain PCA takes each sample into account equally, which is harmful for the representation of the template.

We also compare our system with other baseline methods as part of an ablation study, from baseline cosine distance \textbf{Cos} to the proposed quality-aware subspace-to-subspace similarity \textbf{QCos+QSub-VPM}. As we gradually modify the method by including quality-aware cosine distance \textbf{QCos}, quality-aware subspace learning \textbf{QSub} and variance-aware projection metric \textbf{VPM}, we can see the performance also gradually improves, especially for IJB-B and IJB-S datasets.

From the results above, we observe the following:
\begin{itemize}
\item The proposed system performs the best in general, which shows the effectiveness of 1) learning subspace as template representation, 2) matching video pairs using the subspace-to-subspace similarity measure and 3) utilizing quality and variance information to compute exemplars, learn subspaces and measure similarity.

\item \textbf{QCos} generally performs better than \textbf{Cos}, which shows that quality-aware exemplars weigh the samples according to their quality and better represent the image sets than plain average exemplars.

\item In most of the cases, \textbf{Cos+Sub-PM} achieve higher performance than \textbf{Cos}. It implies that subspace can utilize the correlation information between samples and is a good complementary representation of exemplars as global information.

\item \textbf{QCos+QSub-PM} performs better than \textbf{QCos+Sub-PM} in general. It shows that similar to \textbf{QCos}, we can learn more representative subspaces based on the quality of samples.

\item \textbf{QCos+QSub-VPM} works better than \textbf{QCos+QSub-PM} in most of the experiments. It implies that by considering the variances of bases in the subspaces, \textbf{VPM} similarity is more robust to variations in the image sets.

\item The improvement of the proposed system over the compared algorithms is consistent under both \textbf{with filtering} and \textbf{without filering} configurations on the IJB-S dataset. It shows that our method is effective for both high-quality and low-quality tracklets in surveillance videos.

\item For IJB-S, the performance on surveillance-to-surveillance protocol is in general lower than the performance on other protocols. This is because the gallery templates of this protocol are from low-quality surveillance videos, while the rest two protocols have galleries from high-resolution still images.

\item The fusion of Network D and E does not perform as well as single Network D on surveillance-to-surveillance protocol, especially at higher rank accuracy. It is probably because of the low-quality galleries in this protocol which Network E cannot represent well. 

\item On IJB-S, the proposed method performs better than state-of-the-art network ArcFace \cite{arcface} in general, especially on surveillance-to-single and surveillance-to-booking protocols, which shows the discriminative power of the features from the proposed networks. ArcFace still performs better on surveillance-to-surveillance protocol. But the results also show that using the quality-aware subspace-to-subspace similarity improves the performance for ArcFace features as well.

\item On MBGC and FOCS, ArcFace performs better in the walking-vs-walking protocol but Network D outperforms ArcFace on more challenging protocols like activity-vs-activity. Also, by applying the proposed subspace-to-subspace similarity on both features, the performance consistently improves, which shows its effectiveness on different datasets and using different features.

\item For the FOCS dataset, the performance of our system surpasses the human performance, which again demonstrates the effectiveness of the proposed system. 
\end{itemize}

\section{Conclusions}\label{sec:conclusion}
In this paper, we proposed an automatic face recognition system for unconstrained video-based face recognition tasks. The proposed system consists of modules for face detection and alignment, face association and tracking, face representation, subspace learning and subspace-to-subspace similarity-based matching. We evaluated our system on four video datasets. The experimental results demonstrate the superior performance of the proposed system.

{\small
\section*{Acknowledgment}

This research is based upon work supported by the Office of the Director of National Intelligence (ODNI), Intelligence Advanced Research Projects Activity (IARPA), via IARPA R\&D Contract No. 2019-022600002. The views and conclusions contained herein are those of the authors and should not be interpreted as necessarily representing the official policies or endorsements, either expressed or implied, of the ODNI, IARPA, or the U.S. Government. The U.S. Government is authorized to reproduce and distribute reprints for Governmental purposes notwithstanding any copyright annotation thereon.

}

\ifCLASSOPTIONcaptionsoff
  \newpage
\fi



%

%

\bibliographystyle{IEEEtran}
\bibliography{refs}






\end{document}